\documentclass[10pt,journal,compsoc]{IEEEtran}

\ifCLASSOPTIONcompsoc
  \usepackage[nocompress]{cite}
\else
  \usepackage{cite}
\fi
\usepackage{amsmath,amssymb,array,booktabs,graphicx,makecell,multirow,tabularx,url}
\usepackage[hidelinks]{hyperref}
\newcommand{\suppsec}[1]{\hyperref[#1]{Appendix~\ref*{#1}}}
\newcommand{\suppfig}[1]{\hyperref[#1]{Fig.~\ref*{#1}}}
\newcommand{\supptab}[1]{\hyperref[#1]{Table~\ref*{#1}}}

\begin{document}

\title{Learning Across Temporal Graph Representation Silos}

\author{Minwoo~Yu and Young-guk~Ha%
\IEEEcompsocitemizethanks{%
\IEEEcompsocthanksitem M. Yu and Y.-g. Ha are with Smart Computing
Laboratory, Department of Computer Science \& Engineering, Konkuk University,
Seoul 05029, Republic of Korea. E-mail: \{snowypainter, ygha\}@konkuk.ac.kr.
\IEEEcompsocthanksitem Y.-g. Ha is the corresponding author.}}

\markboth{IEEE Transactions on Knowledge and Data Engineering}%
{Yu and Ha: Learning Across Temporal Graph Representation Silos}

\maketitle
\begin{abstract}
Temporal graph data are recorded in different representations, but whether these differences also determine the scope of learned knowledge reuse remains unclear. This study investigates whether predictive structure learned from relational histories can be reused between continuous-time dynamic graphs (CTDGs) and temporal knowledge graphs (TKGs). Bidirectional reuse emerges while retaining distinct input and prediction interfaces and independent vocabularies, and sources with the same representation are not always more useful. Controlled reorganizations of source histories show that the organization contributing to reuse varies across source--target conditions. We show that learned aspects of Relational Dynamics, the predictive structure present in relational histories, can be reused across conventional representation boundaries. Independently pretrained TKG and CTDG learners can also be fused at the score level to improve prediction on a third dataset without entity or relation-semantic alignment. Two proofs of concept in recommendation and security detection illustrate that external temporal sources without direct alignment can also become useful pretraining resources.
\end{abstract}


\IEEEpeerreviewmaketitle
\bstctlcite{BSTcontrol}

\section{Introduction}
\label{sec:introduction}

Foundation models require reusable knowledge. The first question in temporal graph learning is therefore not which architecture to scale up, but what learned from different temporal graph datasets can be reused across representation boundaries. This study asks whether knowledge learned from recurring relations and patterns of change over time remains reusable when the representation in which the data are recorded changes.

Temporal graph data are recorded in different representations. Continuous-time dynamic graphs (CTDGs) describe timestamped interaction streams, whereas temporal knowledge graphs (TKGs) describe typed facts of the form \texttt{(subject, relation, object, time)}~\cite{kazemi2020survey}. They differ in input structure, vocabulary, queries, and prediction interfaces. Accordingly, recent pretraining and transfer studies have largely expanded the scope of learning within a single representation. For CTDGs, TGPM~\cite{ma2026tgpm} and DyGFM~\cite{yuan2026dygfm} address pretraining and adaptation across dynamic graph domains. For TKGs, POSTRA~\cite{pan2025postra} and FITTER~\cite{pan2026fitter} study transfer to new entity and relation vocabularies and graphs. Existing approaches thus generally choose a representation first and then seek reusable knowledge within it.

Nevertheless, similar predictive regularities recur across different representations. In CTDGs, EdgeBank showed that the recurrence of past interactions provides a strong predictive signal~\cite{poursafaei2022evaluation}; in TKGs, \emph{History Repeats Itself} also reported the predictive power of recurring facts~\cite{gastinger2024history}. A more fundamental question is therefore whether predictive structure exists that is actually reusable across heterogeneous temporal graph representations.

Section~\ref{subsec:existence} examines this question through frozen cross-representation transfer between CTDGs and TKGs. We retain separate input interfaces and prediction heads for each representation and transfer only the history encoder learned on the source to the target. Despite sharing neither entity vocabularies nor relation semantics, source-pretrained backbones provide greater utility than random backbones in both directions, and such reuse is also observed across multiple backbones. Thus, learned predictive structure can survive changes in representation and prediction interface.

Does this result simply mean that a pretrained network is better than a random network? Section~\ref{subsec:organization} shows that downstream utility changes when source event--time assignments and partner organization are perturbed before transfer. In particular, even partial partner corruption that preserves the source training budget and similar source validation performance yields lower target AP and AUC than the intact source in all three seeds for ICEWS14 \(\rightarrow\) Wikipedia. Transfer utility therefore depends not only on whether pretraining occurred, but also on what was learned from the source relational history. 

Relational Dynamics denotes predictive structure present in relational histories. A model captures part of this structure through learning, which we call Learned Relational Dynamics. The part that is also useful for prediction in other datasets or representations is termed Reusable Relational Dynamics. Reusability depends not only on the structure learned from the source, but also on the target and task to which it is applied.

Section~\ref{subsec:functional_profiles} makes this distinction concrete through controlled reorganizations of source histories. We alter partners, event sequence, temporal placement, or relation direction while retaining source competence and then measure changes in target utility. Some conditions favor the original event sequence, whereas others transfer better after source histories are reorganized. Learned Relational Dynamics are therefore not reused in their entirety; the useful part varies with the target and task.

Are these relational dynamics specific to a particular dataset pair? Section~\ref{subsec:scope} evaluates multiple CTDGs and TKGs, source--target pairs in both directions, and target-data regimes of 5\%, 20\%, and 100\%. Cross-representation reuse recurs in most conditions, and source initialization also provides gains under fine-tuning. Reusable predictive knowledge can thus be accumulated from heterogeneous temporal graph data.

This raises a further question for foundation models: is it sufficient to organize reusable knowledge within representation-specific silos such as CTDGs and TKGs? Section~\ref{subsec:utility} compares all 30 directed transfers among six datasets, excluding self-transfer. Source utility is not consistently separated by representation membership, and cross-representation sources repeatedly prove more useful than same-representation sources for the same target. Representation identity does not fully explain which sources provide reusable knowledge.

\begin{figure*}[t]
\centering
\includegraphics[width=\linewidth]{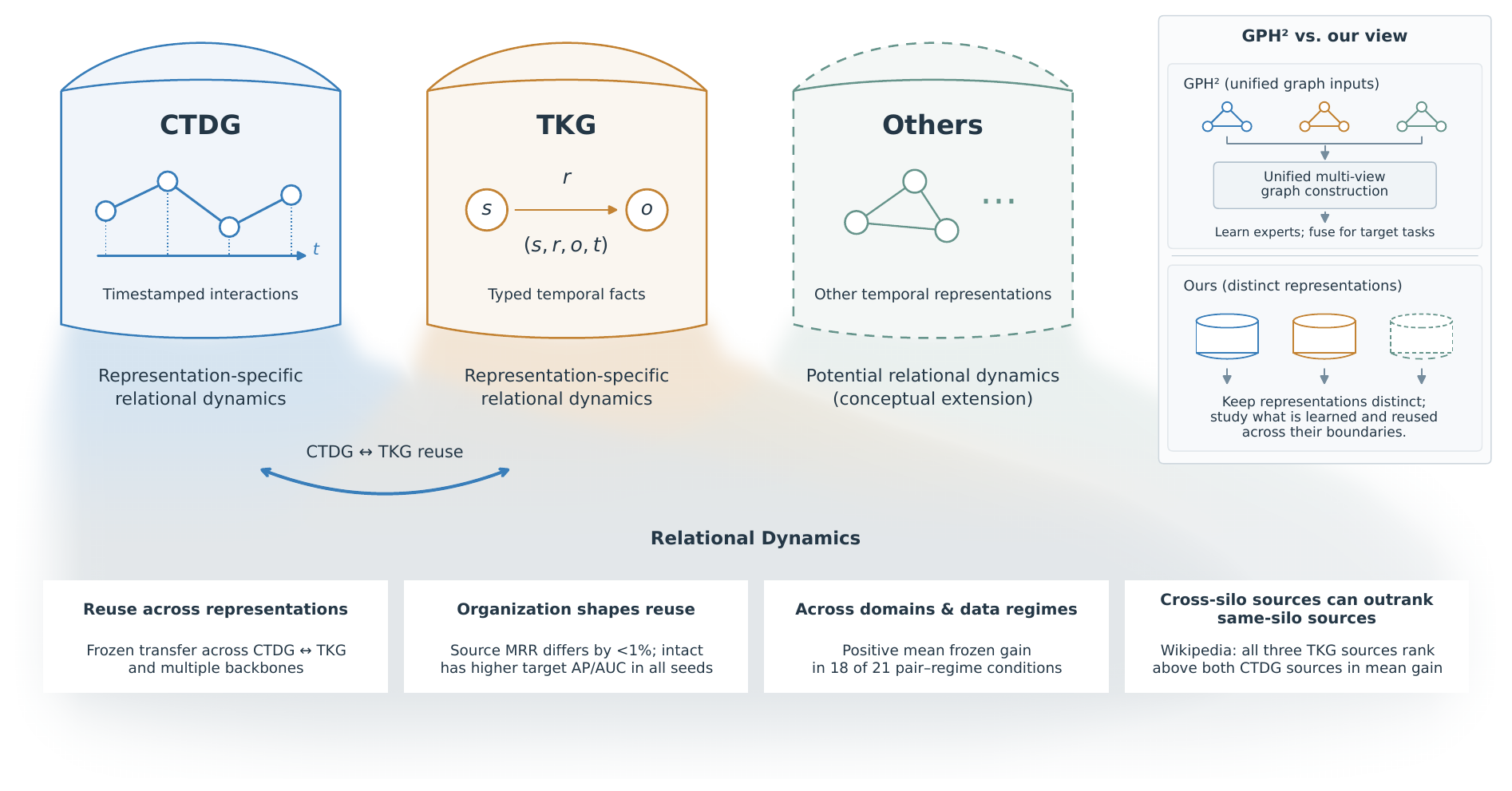}
\caption{\textbf{Relational dynamics across temporal graph representation silos.} We ask what can be learned and reused beneath different observation and prediction formats. Representation-specific colors transitioning to a neutral color below illustrate the perspective of relational dynamics as a common object of study. Color boundaries and overlaps are conceptual and assume neither identical dynamics nor a common latent space. The dashed ``Others'' silo denotes future extensions. The bottom summarizes the existence of reuse, dependence on source organization, reuse across domains and training fractions, and source utility rankings. The right contrasts GPH\textsuperscript{2}~\cite{liang2026gph2}, which designs unified multi-view inputs and expert fusion, with our focus on the object of learning itself across representations.}
\label{fig:silos}
\end{figure*}

Figure~\ref{fig:silos} summarizes the perspective on temporal graph foundation models suggested by these results. CTDGs and TKGs need not be converted into a single representation. Their observation and prediction interfaces can remain distinct while sharing a learner that processes temporal-relational histories beneath them.

We establish three findings. First, CTDG\(\leftrightarrow\)TKG transfer demonstrates that aspects of Relational Dynamics remain reusable after the representation and prediction interface change. Second, source interventions and functional characterization show that reuse depends on source relational organization and that the contributing organization varies across source--target conditions. Third, experiments across domains and data regimes, together with a complete transfer matrix, show that source utility orderings cross conventional representation taxonomies. Restricting foundation-model pretraining sources to representation-specific silos can therefore exclude more useful knowledge. Section~\ref{subsec:fusion} further tests whether independently pretrained TKG and CTDG learners can be used jointly on a third target. Their score-level combination demonstrates that reused predictive knowledge can provide additional utility without directly aligning the sources. Finally, two proofs of concept (PoCs) in recommendation and security detection illustrate how this reuse can turn external temporal sources without direct alignment into practical learning resources.

\section{Related Work}
\label{sec:related_work}

\paragraph{Temporal graph representations and learning.}
Kazemi et al.~\cite{kazemi2020survey} distinguish the temporal representations of CTDGs and discrete-time dynamic graphs, cover dynamic KGs, and organize models from an encoder--decoder perspective. Temporal representation and relation typing are distinct axes. In this paper, CTDG denotes data represented as interaction streams, whereas TKG denotes data represented as typed temporal facts; we do not assume that they are mutually exclusive graph classes. Rather than using this distinction as a model taxonomy, we study whether learned history-processing capabilities are reusable across temporal graph representations.

UTG compares DTDG and CTDG models within a common framework through snapshot--event conversion and unified training and evaluation procedures~\cite{huang2025utg}. Our study tests the reuse of source-learned predictive structure between independent CTDGs and TKGs, providing empirical evidence for relational dynamics.

\paragraph{Relational regularities in temporal prediction.}
By remembering past edges, EdgeBank demonstrated the importance of recurrence and evaluation conditions in dynamic link prediction~\cite{poursafaei2022evaluation}. \emph{History Repeats Itself} analyzed the predictive power of baselines that exploit recurring facts in TKG forecasting~\cite{gastinger2024history}. These findings show that similar predictive regularities arise independently in different representations. However, both studies conduct their analyses within their respective representations and do not directly test whether such regularities can be learned and reused across representations.

\paragraph{Heterogeneous graph integration and pre-training.}
AdaVQGP connects graphs from different domains through a shared semantic codebook and virtual semantic nodes, combining feature-level alignment with model-level structural transfer~\cite{xie2026adavqgp}. GPH\textsuperscript{2} constructs unified multi-view graphs from homogeneous and heterogeneous graphs, independently pretrains an expert for each graph, and freezes, aligns, and fuses these experts downstream~\cite{liang2026gph2}. In recommendation, IGSRec refines an LLM-based semantic graph using observed user--item interactions and combines embeddings learned from the interaction and semantic graphs~\cite{jeong2025interaction}.

\paragraph{Temporal patterns and cross-domain learning.}
ReMIT combines temporal-, neighborhood-, community-, and graph-level information through feature pretraining and knowledge distillation for temporal graph learning~\cite{liu2025remit}. Our study asks whether predictive structure learned from relational histories remains reusable across distinct temporal graph representations and prediction interfaces.

TGPM represents timestamped interaction sequences as interaction patches derived from temporal random walks and learns generalized evolving patterns through masked token modeling and next-time prediction~\cite{ma2026tgpm}. It evaluates cross-domain transfer of the learned model on temporal link prediction, including ICEWS1819~\cite{ma2026tgpm}.

DyGFM proposes learning and adaptation that account for semantic and temporal differences across dynamic graph domains~\cite{yuan2026dygfm}. FLIP reduces dependence on graph-specific vocabularies through factorized edge-event intensity and addresses transfer between CTDGs by reusing parameters without target gradient updates while constructing states from target histories~\cite{yu2026flip}. In this line of work, domains and node vocabularies change, but the interaction-based CTDG representation is retained.

For TKGs, POSTRA studies fully inductive temporal reasoning and transfer to graphs with new entities, relations, and timestamps~\cite{pan2025postra}. FITTER transfers across TKG domains through vocabulary-agnostic structural learning~\cite{pan2026fitter}. Here too, domains and vocabularies change, but the typed temporal fact representation is retained.

STKGN combines a spatial-temporal knowledge graph with a continuous-time dynamic graph network for event prediction~\cite{huai2023stkgn}. This approach combines graphs and modeling components within one application, whereas our setting reuses a learned history learner between independent CTDGs and TKGs while changing the input and prediction interfaces.

\paragraph{Multi-source prediction fusion.}
In multi-source domain adaptive object detection, Ma et al.~\cite{ma2026independent} use feature alignment and cross-domain information exchange and combine predictions from multiple adapted detectors with weights. We investigate whether learners independently pretrained on distinct TKG and CTDG sources can be transferred to a third target and used jointly without cross-source entity or relation-semantic alignment.

\paragraph{Positioning of this study.}
To our knowledge, this is the first empirical study of what is reusable beneath distinct temporal graph representations themselves. Rather than designing a common formulation for heterogeneous graphs, we preserve representation-specific interfaces and test whether learned predictive structure transfers across them. We examine the reusability of learned aspects of Relational Dynamics, their dependence on source organization, and their transfer scope.

\section{Problem Formulation}
\label{sec:problem_definition}

A temporal graph dataset \(D\) contains relational histories recorded in a representation \(\mathsf R(D)\). A query \(q\) at time \(t\) is predicted using the preceding observation history \(\mathcal H_D(t)\). We study whether predictive structure learned from one dataset's history can be reused in another while retaining the inputs and prediction tasks of their respective representations. Although this formulation is not restricted to CTDGs and TKGs, our empirical study focuses on these two representations, for which substantial research on models and prediction tasks is available.

\paragraph{Conceptual distinction.}
Relational Dynamics is the predictive structure present in relational histories. Learned Relational Dynamics is the part captured by a model during learning, and Reusable Relational Dynamics is the part that remains useful for target prediction in another dataset or representation. Reusable Relational Dynamics is therefore not a fixed property of a source encoder alone, but a property of the transfer relation jointly determined by source, target, and prediction query.

\paragraph{Terminology and transfer direction.}
The source \(S\) is the dataset on which transferable parameters are first learned, and the target \(T\) is the dataset that receives these parameters for adaptation and predictive evaluation. \(S\rightarrow T\) denotes this direction of learned information transfer. For example, in Wikipedia \(\rightarrow\) ICEWS14, Wikipedia is the source and ICEWS14 is the target. Domain denotes the application environment in which data are observed, whereas representation denotes the format used to record events and relations. We use datasets as the unit of domain comparison; same- and cross-representation indicate whether the source and target share the same CTDG/TKG designation.

A dataset-specific input interface \(A_{\phi_D}\) transforms observations into encoder inputs, a shared history encoder \(F_\theta\) (backbone) processes relational histories, and a prediction head \(G_{\psi_D}\) transforms its outputs into prediction scores for queries. Shared parameters are transferred or shared across datasets, whereas private parameters are learned separately for each dataset. The prediction function is
\begin{equation}
\label{eq:prediction}
s_D(q)=G_{\psi_D}\!\left(q,F_\theta\!\left(A_{\phi_D}(q,\mathcal H_D(t))\right)\right).
\end{equation}

Our implementation and experiments instantiate this formulation through CTDG interaction \((u,v,t)\) prediction and TKG head/tail entity forecasting. Here, \(q\) is a pairwise query that includes the candidate being evaluated, and \(A_{\phi_D}\) extracts and processes the strict-past histories \(\mathcal H_u^D(t),\mathcal H_v^D(t)\) incident to the two endpoints from \(\mathcal H_D(t)\). We therefore distinguish the general interface separation in Eq.~\eqref{eq:prediction} from the two-endpoint implementation used here. \(F_\theta\) learns part of the Relational Dynamics in relational histories; this study examines the part reused for target prediction. Section~\ref{subsec:existence} evaluates reuse through frozen transfer, Section~\ref{subsec:organization} evaluates sensitivity to source organization, and Section~\ref{subsec:functional_profiles} distinguishes which source-history organization contributes to reuse for each target.

\paragraph{Cross-representation reuse.}
When the source \(S\) and target \(T\) satisfy \(\mathsf R(S)\ne\mathsf R(T)\), only the shared parameters \(\theta_S\) learned on the source are transferred to the target. Let \(M_T^{\mathrm{frz}}(\theta)\) denote the evaluation score obtained after learning target-private parameters \((\phi_T,\psi_T)\) with \(\theta\) fixed. The reuse gain is
\begin{equation}
\label{eq:reuse_gain}
\Delta_{S\rightarrow T}=M_T^{\mathrm{frz}}(\theta_S)-M_T^{\mathrm{frz}}(\theta_0),
\end{equation}
where \(\theta_0\) is a randomly initialized encoder. Under identical target training conditions, \(\Delta_{S\rightarrow T}>0\) indicates a gain from reusing the source-learned encoder. Both conditions learn private parameters from the target training data.

\paragraph{Functional characterization.}
A positive reuse gain shows that source-learned structure is useful for the target. To characterize that structure, we define a controlled operator \(C_k\) that reorganizes one aspect \(k\) of the source relational history. The variant \(D_S^{(k)}\) preserves the event count and specified statistics of the original source while reorganizing partners, event sequence, temporal placement, or relation direction:
\begin{equation}
D_S^{(k)}=C_k(D_S),\qquad
\theta_S=\mathcal A(D_S),\qquad
\theta_S^{(k)}=\mathcal A(D_S^{(k)}).
\end{equation}
The two encoders use identical initialization, architecture, optimizer, and update count. Before comparing target utility, we measure their relative performance difference on the same original source history:
\begin{equation}
\label{eq:source_gap}
g_S^{(k)}=
\frac{\left|M_S^{\mathrm{orig}}(\theta_S)-M_S^{\mathrm{orig}}(\theta_S^{(k)})\right|}
{M_S^{\mathrm{orig}}(\theta_S)}.
\end{equation}
Only conditions satisfying \(g_S^{(k)}\leq0.03\) for all three seeds enter the target comparison. For source-competence-matched conditions, the target contribution of history aspect \(k\) is
\begin{equation}
\label{eq:functional_contribution}
B_{S\rightarrow T}^{(k)}=
M_T^{\mathrm{frz}}(\theta_S)-M_T^{\mathrm{frz}}(\theta_S^{(k)}).
\end{equation}
Here \(B>0\) indicates that the original organization contributes to target reuse, whereas \(B<0\) indicates better transfer from the reorganized source. Because the same organization may function differently across targets and queries, we interpret each \(S\times T\times k\) condition as a functional profile.

\section{Experimental Instantiation}
\label{sec:method}

To empirically examine cross-representation predictive reuse, we instantiate Eq.~\eqref{eq:prediction} using representation-specific input and prediction interfaces and a transferable history encoder. We adopt DyGFormer's history-processing components as the primary experimental backbone~\cite{yu2023dygformer}.

\subsection{Temporal-Relational Inputs}
\label{subsec:inputs}
For each endpoint, we use up to 63 most recent strict-past events and a central-node token, fixing the input length to 64 including padding. Each position comprises four channels: node, event, relative time, and neighbor co-occurrence. The node channel uses raw features and is set to a zero vector for TKGs. We do not use trainable entity-ID embeddings; IDs are used for history lookup and co-occurrence computation. The time channel encodes the time difference \(t-\tau_i\), and the co-occurrence channel encodes the number of times the corresponding neighbor appears in each endpoint's input sequence.

The event channel uses edge features for CTDGs and the sum of embeddings for the historical relation \(\rho_i\) and its direction \(d_i\) relative to the central node for TKGs:
\begin{equation}
\label{eq:event}
e_i^D=
\begin{cases}
x_i^{\mathrm{edge}}, & D\text{ is CTDG},\\
E_{\mathrm{rel}}^D(\rho_i)+E_{\mathrm{dir}}^D(d_i), & D\text{ is TKG}.
\end{cases}
\end{equation}
The direction \(d_i\) distinguishes outgoing, incoming, and self-loop events, preserving fact direction in both endpoints' histories. Relation and direction embeddings and the event-channel projection are dataset-specific parameters.

\subsection{Shared History Encoder and Prediction}
\label{subsec:encoder}
Each of the four channels is grouped into patches of length 2 and projected, after which the projected channels are concatenated. Let \(Z_a\) denote the patch sequence for endpoint \(a\). The two sequences are processed jointly and pooled separately for each endpoint:
\begin{equation}
\label{eq:encoder}
\begin{aligned}
\relax[Y_u;Y_v]&=\operatorname{Transformer}_\theta([Z_u;Z_v]),\\
h_a&=P_O\!\left(\operatorname{MeanPool}(Y_a)\right),\quad a\in\{u,v\}.
\end{aligned}
\end{equation}
Joint self-attention makes \(h_u,h_v\) pair-conditioned representations that reflect both histories. The implementation uses 50-dimensional projections per channel, a two-layer, two-head Transformer, and 172-dimensional output vectors. The CTDG head scores the two vectors, whereas the TKG head additionally receives the query relation embedding:
\begin{equation}
\label{eq:heads}
\begin{aligned}
s_{\mathrm{CTDG}}(u,v,t)&=G_{\psi_C}([h_u\Vert h_v]),\\
s_{\mathrm{TKG}}(u,\rho,v,t)&=G_{\psi_K}([h_u\Vert h_v\Vert E_{\mathrm{rel}}^D(\rho)]).
\end{aligned}
\end{equation}
Each head is an MLP with a hidden layer and ReLU that outputs a single logit. In TKGs, histories and queries use the same relation embedding table; evaluation replaces either the head or tail candidate while retaining the original fact direction.

\subsection{Learning and Parameter Reuse}
\label{subsec:learning}
We construct one negative per positive and apply binary cross-entropy to their scores \(s^+,s^-\):
\begin{equation}
\label{eq:loss}
\mathcal L_D=-\tfrac12\bigl[\log\sigma(s^+)+\log\sigma(-s^-)\bigr].
\end{equation}
CTDG training replaces the destination; TKG training replaces the entity in the prediction direction and uses both head and tail queries for each fact. Negative sampling and evaluation candidate sets follow each target's protocol.

The shared parameters \(\theta\) comprise the time and co-occurrence encoders, node, time, and co-occurrence projections, the Transformer, and the output projection. The private parameters \((\phi_D,\psi_D)\) comprise the event projection, TKG relation and direction embeddings, and the prediction head. Only \(\theta\) is transferred from the source; target-private parameters are freshly initialized, and feature tables and histories come from the target.

Frozen transfer fixes \(\theta_S\) and learns only target-private parameters. Random-frozen is a control that fixes random parameters \(\theta_0\). Source-finetune starts from \(\theta_S\) and updates the entire model, whereas target-only trains the entire model from random initialization. We match target-private initialization and training budgets between paired conditions. Frozen comparisons measure reuse of the learned encoder, while fine-tuning comparisons measure the effect of source initialization.

\section{Experiments}
\label{sec:experiments}

\subsection{Experimental Setup}
\label{subsec:setup}
We use DyGFormer as the primary backbone and additionally examine frozen reuse with GraphMixer and TGAT. Frozen comparisons measure source-frozen $-$ random-frozen, and fine-tuning comparisons measure source-finetune $-$ target-only, with matched target training budgets and private initialization. We report means, standard deviations, and paired gains across seeds 7, 17, and 37. Table~\ref{tab:setup} summarizes the datasets and evaluation settings.

\begin{table}[t]
\caption{Datasets and evaluation settings.}
\label{tab:setup}
\scriptsize\renewcommand{\arraystretch}{1.1}
\setlength{\tabcolsep}{2pt}
\begin{tabularx}{\columnwidth}{@{}lXX@{}}
\toprule
Setting & CTDG & TKG \\
\midrule
Dataset & Wikipedia, Reddit, MOOC & ICEWS14, GDELT, ICEWS05-15 \\
Task / metrics & Link prediction; AP, AUC & Head/tail forecasting; filtered MRR, Hits@1/3/10 \\
Checkpoint selection & Full-validation AP & 100-negative MRR; full validation (256 queries for the DyGFormer Wiki \(\rightarrow\) ICEWS14 diagnostic) \\
Test evaluation & 1 random destination per positive & Bidirectional queries \(\times\) all entities; timestamp filtering \\
\bottomrule
\end{tabularx}
\end{table}

The GraphMixer and TGAT experiments and Sections~\ref{subsec:scope} and~\ref{subsec:utility} use the first 32,768 events of each dataset, split chronologically into approximately 70/15/15, with 10 training epochs each for the source and target. The 5\% and 20\% target training conditions retain only the most recent training events, reducing supervision and training history together. Validation and test sets remain fixed, and evaluation histories contain only retained training events and validation/test observations preceding the query. Parameters are not updated during evaluation.

The bidirectional DyGFormer existence experiment evaluates the full test sets of the original data splits. GraphMixer, TGAT, and the complete source--target matrix use frozen transfer with 100\% target training data. \suppsec{sec:appendix_details} provides experiment-specific details, including negative sampling, history updates, and corruption controls.

The recommendation PoC in Section~\ref{subsec:recommendation} transfers an ICEWS14-pretrained encoder to LastFM and fine-tunes the entire model, comparing it with target-only training of the same model. The first 32,768 LastFM interactions are chronologically split into 22,937/4,915/4,916 training/validation/test interactions. We use the most recent 5\% and 20\% of training interactions (1,147 and 4,588). Target training uses Adam ($10^{-4}$), a batch size of 512, and 10 epochs. Checkpoints are selected by 100-negative NDCG@10 on 256 fixed validation queries, and Recall@10 and NDCG@10 are evaluated over all 1,000 items on a separate set of 256 fixed test queries. Previously listened-to items remain candidates, and ties receive average ranks. History updates follow the procedure above; no item--entity alignment is used.

\subsection{Representation Boundaries Do Not Bound Predictive Reuse}
\label{subsec:existence}
Predictive structure learned in one representation remains useful after the representation and prediction interface change. Table~\ref{tab:existence} shows bidirectional frozen reuse between the Wikipedia CTDG and ICEWS14 TKG.

\begin{table*}[t]
\caption{Bidirectional frozen transfer with DyGFormer.}
\label{tab:existence}
\centering\scriptsize\setlength{\tabcolsep}{3pt}
\begin{tabular}{@{}lllrrrrr@{}}
\toprule
Direction & Metric & Frozen backbone & Seed 7 & Seed 17 & Seed 37 & Mean $\pm$ SD & Mean $\Delta$ \\
\midrule
Wiki $\rightarrow$ ICEWS14 & Test MRR & Random & 0.1225 & 0.1177 & 0.1092 & $0.1165\pm0.0067$ & --- \\
Wiki $\rightarrow$ ICEWS14 & Test MRR & Source-pretrained & 0.2168 & 0.1894 & 0.2115 & $\mathbf{0.2059\pm0.0145}$ & $\mathbf{+0.0894}$ \\
ICEWS14 $\rightarrow$ Wiki & Test AP & Random & 0.8414 & 0.8264 & 0.8713 & $0.8463\pm0.0228$ & --- \\
ICEWS14 $\rightarrow$ Wiki & Test AP & Source-pretrained & 0.9793 & 0.9808 & 0.9865 & $\mathbf{0.9822\pm0.0038}$ & $\mathbf{+0.1359}$ \\
ICEWS14 $\rightarrow$ Wiki & Test AUC & Random & 0.8320 & 0.8213 & 0.8601 & $0.8378\pm0.0201$ & --- \\
ICEWS14 $\rightarrow$ Wiki & Test AUC & Source-pretrained & 0.9763 & 0.9789 & 0.9845 & $\mathbf{0.9799\pm0.0042}$ & $\mathbf{+0.1421}$ \\
\bottomrule
\end{tabular}
\end{table*}

The source-pretrained backbone outperforms the random backbone in all seeds in both directions. Test MRR increases from 0.1165 to 0.2059 for Wiki $\rightarrow$ ICEWS14, and test AP increases from 0.8463 to 0.9822 for ICEWS14 $\rightarrow$ Wiki. These improvements occur even with the backbone fixed during target training, demonstrating that source-learned predictive structure is reusable across representations and prediction interfaces.

\begin{figure*}[t]
\centering
\includegraphics[width=\linewidth]{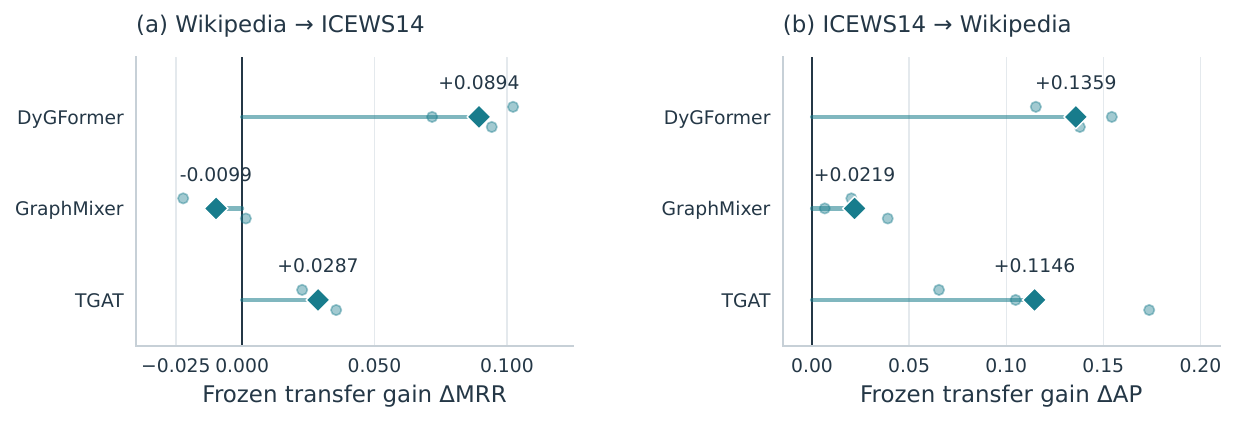}
\caption{\textbf{Frozen reuse across representations and backbones.} Each point represents a test gain of source-frozen $-$ random-frozen; large diamonds denote three-seed means and small points individual seeds. Values to the right of zero indicate reuse gains. (a) MRR for Wikipedia $\rightarrow$ ICEWS14. (b) AP for ICEWS14 $\rightarrow$ Wikipedia. DyGFormer uses the original data splits, while GraphMixer and TGAT use 32,768-event splits; comparisons therefore concern paired gains within each backbone.}
\label{fig:backbones}
\end{figure*}

\paragraph{Across backbones.}
Cross-representation reuse is not confined to DyGFormer. Figure~\ref{fig:backbones} shows reuse gains by transfer direction for DyGFormer, the MLP-mixer-based GraphMixer, and the temporal-graph-attention-based TGAT. \supptab{tab:backbones} in \suppsec{sec:appendix_backbones} provides the scores and paired gains for each backbone.

TGAT yields positive gains in all seeds in both directions, and GraphMixer does so in all seeds for ICEWS14 $\rightarrow$ Wiki. GraphMixer's mean gain for Wiki $\rightarrow$ ICEWS14 is negative, indicating that the utility of reuse depends on the backbone and transfer direction. Improvements recurring across different encoders show that representation differences do not block reusable predictive structure and that this phenomenon is not specific to one architecture.

\paragraph{Same-dataset representation diagnostic.}
To complement transfer across datasets, \supptab{tab:same_dataset} in \suppsec{sec:appendix_same_dataset} compares pretraining with relation information retained or hidden in the same ICEWS14 source history. Pretraining on the relation-hidden CTDG-style view raises test MRR by 0.0746 over training from scratch; retaining relations in the TKG view yields a further 0.0213. Thus, history-based predictive structure learned without relation typing can provide useful initialization for subsequent TKG forecasting. Pretraining and adaptation use nonoverlapping training events, but additional optimization relative to scratch and the information difference introduced by hiding relations remain. This diagnostic therefore provides supporting evidence of reuse, rather than identifying an independent causal effect of representation change.

\subsection{Cross-Representation Reuse Depends on Source Organization}
\label{subsec:organization}
The utility of predictive structure reused across representations depends on the organization of the source from which it was learned. Table~\ref{tab:corruption} compares transfer utility after perturbing source event--time assignments (time shuffle) or interaction partners (partner shuffle).

\begin{table*}[t]
\caption{Transfer performance under source-organization corruption. $\Delta$ denotes Intact $-$ Corrupted. Wiki $\rightarrow$ ICEWS14 uses 256 validation queries; the reverse direction uses the full test set.}
\label{tab:corruption}
\centering\small\setlength{\tabcolsep}{4pt}
\begin{tabular}{@{}lrrrrrr@{}}
\toprule
Direction / metric & Seed & Intact & Time shuffle & Partner shuffle & $\Delta$ Time & $\Delta$ Partner \\
\midrule
Wiki $\rightarrow$ ICEWS14 · Val. MRR & 7 & 0.2018 & 0.1425 & 0.1989 & $+0.0593$ & $+0.0028$ \\
Wiki $\rightarrow$ ICEWS14 · Val. MRR & 17 & 0.1899 & 0.2481 & 0.1177 & $-0.0582$ & $+0.0722$ \\
Wiki $\rightarrow$ ICEWS14 · Val. MRR & 37 & 0.2490 & 0.2292 & 0.2550 & $+0.0198$ & $-0.0060$ \\
\textbf{Wiki $\rightarrow$ ICEWS14 · Val. MRR} & \textbf{Mean} & \textbf{0.2136} & \textbf{0.2066} & \textbf{0.1906} & $\mathbf{+0.0070}$ & $\mathbf{+0.0230}$ \\
\midrule
ICEWS14 $\rightarrow$ Wiki · Test AP & 7 & 0.9822 & 0.9834 & 0.9593 & $-0.0012$ & $+0.0229$ \\
ICEWS14 $\rightarrow$ Wiki · Test AP & 17 & 0.9821 & 0.9846 & 0.9658 & $-0.0025$ & $+0.0163$ \\
ICEWS14 $\rightarrow$ Wiki · Test AP & 37 & 0.9867 & 0.9855 & 0.9779 & $+0.0012$ & $+0.0088$ \\
\textbf{ICEWS14 $\rightarrow$ Wiki · Test AP} & \textbf{Mean} & \textbf{0.9837} & \textbf{0.9845} & \textbf{0.9677} & $\mathbf{-0.0008}$ & $\mathbf{+0.0160}$ \\
\bottomrule
\end{tabular}
\end{table*}

Partner shuffle changes cross-representation utility in both directions and reduces performance in all three seeds for ICEWS14 $\rightarrow$ Wikipedia. On average, it reduces MRR by 0.0230 for Wiki $\rightarrow$ ICEWS14 and AP by 0.0160 for ICEWS14 $\rightarrow$ Wikipedia. For Wiki $\rightarrow$ ICEWS14, performance decreases in two of the three seeds but increases by 0.0060 in one, so the effect varies across seeds and directions.

The effect of temporal organization is less uniform. For Wiki $\rightarrow$ ICEWS14, time shuffle reduces performance in two of three seeds, decreasing mean MRR by 0.0070. For ICEWS14 $\rightarrow$ Wikipedia, however, mean AP increases by 0.0008. These experiments therefore do not establish temporal ordering itself as a consistent transfer signal in every direction.

The effect of partner organization remains under equal training budgets and similar source performance. To examine whether poorer source-task performance alone explains this effect, we compare partial partner corruption for ICEWS14 $\rightarrow$ Wikipedia. The 50\% nominal corruption level, selected using source scores alone, changes approximately 35\% of actual destinations, yet source MRR differs by less than 1\% in all three seeds (Table~\ref{tab:matched}).

\begin{table}[t]
\caption{Partial partner-corruption comparison at similar source performance.}
\label{tab:matched}
\centering\scriptsize\setlength{\tabcolsep}{2pt}
\begin{tabular}{@{}llrrr@{}}
\toprule
Seed & Source condition & Source MRR & Target AP & Target AUC \\
\midrule
7 & Intact & 0.749487 & 0.982154 & 0.979528 \\
7 & 50\% partial partner & 0.753258 & 0.981049 & 0.977285 \\
17 & Intact & 0.755403 & 0.982111 & 0.979706 \\
17 & 50\% partial partner & 0.753498 & 0.981675 & 0.978459 \\
37 & Intact & 0.760991 & 0.986747 & 0.985207 \\
37 & 50\% partial partner & 0.753794 & 0.985014 & 0.983014 \\
\bottomrule
\end{tabular}
\end{table}

\begin{figure*}[t]
\centering
\includegraphics[width=\linewidth]{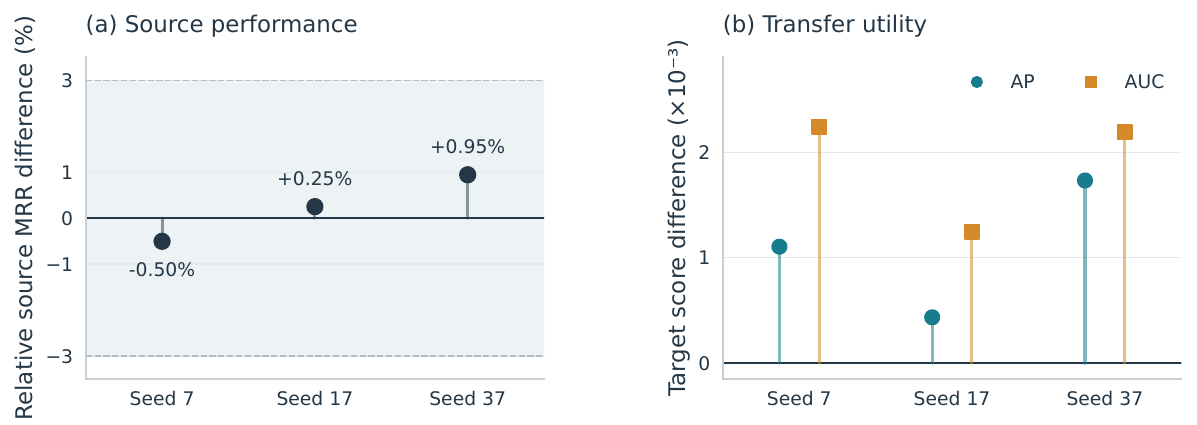}
\caption{\textbf{Different cross-representation utility despite similar source performance.} Both source conditions for ICEWS14 $\rightarrow$ Wikipedia use 10 epochs and 4,980 updates. (a) Relative MRR difference from the intact source, $100\times(\mathrm{Intact}-\mathrm{Partial})/\mathrm{Intact}$; shading denotes the $\pm3\%$ tolerance used for source-only strength selection. Actual differences are below 1\% in all three seeds. (b) Within-seed target AP and AUC differences (Intact $-$ Partial), in units of $10^{-3}$; the intact source is better in all three seeds. The nominal 50\% corruption level was selected using source scores alone and changed approximately 35\% of actual destinations.}
\label{fig:organization}
\end{figure*}

The intact source yields higher target AP and AUC in all three seeds. The paired gains (Intact $-$ Partial) are $+0.001091\pm0.000649$ for AP and $+0.001894\pm0.000561$ for AUC, where $\pm$ denotes the sample standard deviation across the three seeds. The effect is small but directionally consistent. In particular, for seed 7, the intact source yields higher target AP and AUC despite the corrupted model having higher source MRR. Figure~\ref{fig:organization} illustrates this contrast: source performance remains similar, whereas target utility favors the intact source in all three seeds.

The evidence for Relational Dynamics therefore extends beyond reuse gains alone. The next section separates aspects of source history to distinguish what contributes to reuse for each target.

\subsection{Functional Profiles of Reusable Relational Dynamics}
\label{subsec:functional_profiles}
The preceding experiments show that source relational organization changes cross-representation utility. We now functionally characterize which organization is reused by a target. We construct controlled data variants that preserve major source statistics while reorganizing partners, event sequence, temporal placement, or relation direction, and train a new encoder in each condition. These variants do not assume a fixed inventory of Relational Dynamics; they probe whether different aspects of relational history contribute to reuse.

To separate changes to the source task itself, we evaluate the original and variant encoders on the same original source history. Of 21 source--variant conditions, 15 satisfy the 3\% source-competence criterion in all three seeds; only these conditions undergo frozen transfer. \(B_{S\rightarrow T}^{(k)}\) is the target-score difference between the original and controlled-variant encoders. \suppsec{sec:appendix_functional_profiles} reports the preserved statistics, source-performance gaps, and all target results.

Seven of the 15 conditions have the same sign across all three seeds. The original organization is consistently better in one condition, whereas the controlled variant is consistently better in six. Table~\ref{tab:functional_profiles} summarizes the six conditions with clear effect sizes. The seventh, temporal reorganization for Wikipedia \(\rightarrow\) Reddit, is also negative in every seed but has a small mean difference of $-0.0004$ AP.

\begin{table}[t]
\caption{Major functional profiles with consistent directions across three seeds. $B$ is the target-score difference between the original and controlled source variants.}
\label{tab:functional_profiles}
\centering\scriptsize\setlength{\tabcolsep}{2pt}
\begin{tabularx}{\columnwidth}{@{}XXrc@{}}
\toprule
Source $\rightarrow$ Target & Reorganized history aspect & Mean $B\pm$ sample SD & Seed signs \\
\midrule
ICEWS14 $\rightarrow$ ICEWS05-15 & Event sequence & $+0.0330\pm0.0389$ MRR & $+\ +\ +$ \\
ICEWS05-15 $\rightarrow$ Wikipedia & Event sequence & $-0.0041\pm0.0017$ AP & $-\ -\ -$ \\
ICEWS05-15 $\rightarrow$ Wikipedia & Temporal placement & $-0.0048\pm0.0023$ AP & $-\ -\ -$ \\
ICEWS05-15 $\rightarrow$ Wikipedia & Relation direction & $-0.0038\pm0.0044$ AP & $-\ -\ -$ \\
MOOC $\rightarrow$ Reddit & Temporal placement & $-0.0081\pm0.0068$ AP & $-\ -\ -$ \\
Reddit $\rightarrow$ Wikipedia & Temporal placement & $-0.0086\pm0.0108$ AP & $-\ -\ -$ \\
\bottomrule
\end{tabularx}
\end{table}

For ICEWS14 \(\rightarrow\) ICEWS05-15, the contribution of the original event sequence recurs across all three seeds. Reorganizing event sequence while preserving direction- and relation-specific pair frequencies and source competence reduces target MRR in every seed. The seed-level \(B\) values are $+0.0061$, $+0.0154$, and $+0.0776$: their magnitudes vary, but their direction is consistent.

Conversely, for ICEWS05-15 \(\rightarrow\) Wikipedia, encoders trained after reorganizing event sequence, temporal placement, or relation direction outperform the original in AP for all three seeds. Learned Relational Dynamics that are useful on the source are therefore not necessarily the same as Reusable Relational Dynamics useful on a target. The same pattern recurs for temporal reorganization in MOOC \(\rightarrow\) Reddit and Reddit \(\rightarrow\) Wikipedia. Thus, the part of source-learned Relational Dynamics that is actually reused varies with the target.

\subsection{Cross-Representation Reuse Generalizes Across Domains and Data Regimes}
\label{subsec:scope}
Relational dynamics reuse recurs across domains and in both directions beyond a single dataset pair. Across seven cross-representation pairs, three target training fractions (5\%, 20\%, and 100\%), and three seeds, mean frozen gains are positive in 18 of the 21 pair--data conditions. Six of the seven pairs improve at all three training fractions, and the three TKG $\rightarrow$ CTDG pairs yield positive gains in all nine conditions. The reuse observed for Wikipedia $\leftrightarrow$ ICEWS14 thus extends to other datasets and target-data regimes.

Substantial improvements occur even with the source-learned backbone fixed. AP increases from 0.7338 to 0.9476 for ICEWS14 $\rightarrow$ Wikipedia at 20\% training data, and MRR increases from 0.0594 to 0.1216 for Wikipedia $\rightarrow$ ICEWS14 at 100\%. Achieving these gains across different vocabularies and prediction interfaces without modifying the backbone during target training demonstrates the reusability of source-learned predictive structure.

\paragraph{Frozen transfer.}
Table~\ref{tab:scope_frozen} compares mean paired gains (source-frozen $-$ random-frozen) across training fractions. \suppsec{sec:appendix_results} provides all scores and standard deviations.

\begin{table}[t]
\caption{Mean frozen gains across target training fractions.}
\label{tab:scope_frozen}
\centering\scriptsize\setlength{\tabcolsep}{1.5pt}
\begin{tabular}{@{}llrrr@{}}
\toprule
Source $\rightarrow$ Target & Metric & 5\% train & 20\% train & 100\% train \\
\midrule
Wikipedia $\rightarrow$ ICEWS14 & MRR & $+0.0494$ & $+0.0675$ & $+0.0622$ \\
Reddit $\rightarrow$ ICEWS14 & MRR & $+0.0559$ & $+0.0471$ & $+0.0496$ \\
MOOC $\rightarrow$ GDELT & MRR & $-0.0409$ & $-0.0413$ & $-0.0006$ \\
ICEWS14 $\rightarrow$ Wikipedia & AP & $+0.2046$ & $+0.2138$ & $+0.1544$ \\
GDELT $\rightarrow$ Reddit & AP & $+0.0553$ & $+0.0691$ & $+0.0771$ \\
Wikipedia $\rightarrow$ ICEWS05-15 & MRR & $+0.0454$ & $+0.0510$ & $+0.0649$ \\
ICEWS05-15 $\rightarrow$ MOOC & AP & $+0.0453$ & $+0.0394$ & $+0.0680$ \\
\bottomrule
\end{tabular}
\end{table}

Reuse also provides practical benefits under fine-tuning. Mean gains over target-only training are positive in 16 of 21 conditions. In particular, for ICEWS14 $\rightarrow$ Wikipedia at 5\%, AP improves from $0.7483$ to $0.9533$, with a paired gain of $+0.2049\pm0.0040$. For Wikipedia $\rightarrow$ ICEWS05-15 at 100\%, MRR increases from $0.1868$ to $0.2383$. Source training thus provides benefits not only as fixed features but also when the entire model is trained on the target.

\paragraph{Fine-tuning.}
Table~\ref{tab:scope_finetune} compares mean paired gains (source-finetune $-$ target-only) across training fractions. \suppsec{sec:appendix_results} provides all scores and standard deviations.

\begin{table}[t]
\caption{Mean fine-tuning gains across target training fractions.}
\label{tab:scope_finetune}
\centering\scriptsize\setlength{\tabcolsep}{1.5pt}
\begin{tabular}{@{}llrrr@{}}
\toprule
Source $\rightarrow$ Target & Metric & 5\% train & 20\% train & 100\% train \\
\midrule
Wikipedia $\rightarrow$ ICEWS14 & MRR & $-0.0189$ & $+0.0280$ & $+0.0165$ \\
Reddit $\rightarrow$ ICEWS14 & MRR & $-0.0024$ & $-0.0002$ & $+0.0371$ \\
MOOC $\rightarrow$ GDELT & MRR & $-0.0201$ & $+0.0230$ & $-0.0036$ \\
ICEWS14 $\rightarrow$ Wikipedia & AP & $+0.2049$ & $+0.0307$ & $+0.0019$ \\
GDELT $\rightarrow$ Reddit & AP & $+0.0500$ & $+0.0443$ & $+0.0095$ \\
Wikipedia $\rightarrow$ ICEWS05-15 & MRR & $+0.0244$ & $+0.0491$ & $+0.0515$ \\
ICEWS05-15 $\rightarrow$ MOOC & AP & $+0.0563$ & $+0.0786$ & $+0.0776$ \\
\bottomrule
\end{tabular}
\end{table}

Cross-representation reuse persists across target training fractions. Six source--target pairs yield positive frozen gains at 5\%, 20\%, and 100\% of target training data, showing that the utility of source-learned predictive structure is not restricted to a particular target-data regime.

Within this scope of reuse, source and adaptation choices still matter. Frozen gains for MOOC $\rightarrow$ GDELT are negative at all three training fractions, and Reddit $\rightarrow$ ICEWS14 at 5\% yields a frozen gain of $+0.0559$ but a fine-tuning gain of $-0.0024$. The utility of a fixed backbone and the gain from fine-tuning over target-only training are distinct outcomes.

Overall, relational dynamics are broadly reusable across representation boundaries, with utility varying by source--target combination and training conditions. The observation that sources crossing the same representation boundary provide different utility raises the question of whether representation membership alone sufficiently distinguishes useful sources. Section~\ref{subsec:utility} tests this directly by connecting both same- and cross-representation sources to each target.

\subsection{Representation Identity Does Not Fully Determine Source Utility}
\label{subsec:utility}
How well does representation identity distinguish sources whose learned knowledge is useful for a target? Across 30 directed transfers among six datasets, we compare the utility of two same-representation and three cross-representation sources for each target.

Cross-representation sources are useful for multiple targets. Mean paired gains are positive in 14 of the 18 cross-representation transfers, and five of the six targets obtain positive reuse from at least one cross-representation source. We go further by comparing the utility ordering of same- and cross-representation sources for each target.

Table~\ref{tab:benchmark} reports the mean frozen paired gain \(\Delta_{S\rightarrow T}\) across three seeds. \emph{Italics} indicate same-representation transfer, and bold denotes the highest gain for each target.

\begin{table*}[t]
\caption{Mean frozen gains in the complete source--target matrix.}
\label{tab:benchmark}
\centering\small\setlength{\tabcolsep}{4pt}
\begin{tabular}{@{}lrrrrrr@{}}
\toprule
Source $\downarrow$ / Target $\rightarrow$ & Wikipedia & Reddit & MOOC & ICEWS14 & GDELT & ICEWS05-15 \\
 & (AP) & (AP) & (AP) & (MRR) & (MRR) & (MRR) \\
\midrule
Wikipedia & --- & \textbf{\textit{+0.0858}} & \textbf{\textit{+0.0722}} & +0.0622 & $-0.0429$ & +0.0649 \\
Reddit & \textit{+0.1187} & --- & \textit{+0.0524} & +0.0496 & $-0.0276$ & +0.0620 \\
MOOC & \textit{+0.0030} & \textit{+0.0096} & --- & $-0.0016$ & $\mathbf{-0.0006}$ & +0.0014 \\
ICEWS14 & +0.1544 & +0.0710 & +0.0689 & --- & \textit{$-$0.0014} & \textbf{\textit{+0.1811}} \\
GDELT & +0.1454 & +0.0771 & +0.0614 & \textbf{\textit{+0.0819}} & --- & \textit{+0.0974} \\
ICEWS05-15 & \textbf{+0.1577} & +0.0760 & +0.0680 & \textit{+0.0812} & \textit{$-$0.0027} & --- \\
\bottomrule
\end{tabular}
\end{table*}

\begin{figure*}[t]
\centering
\includegraphics[width=\linewidth]{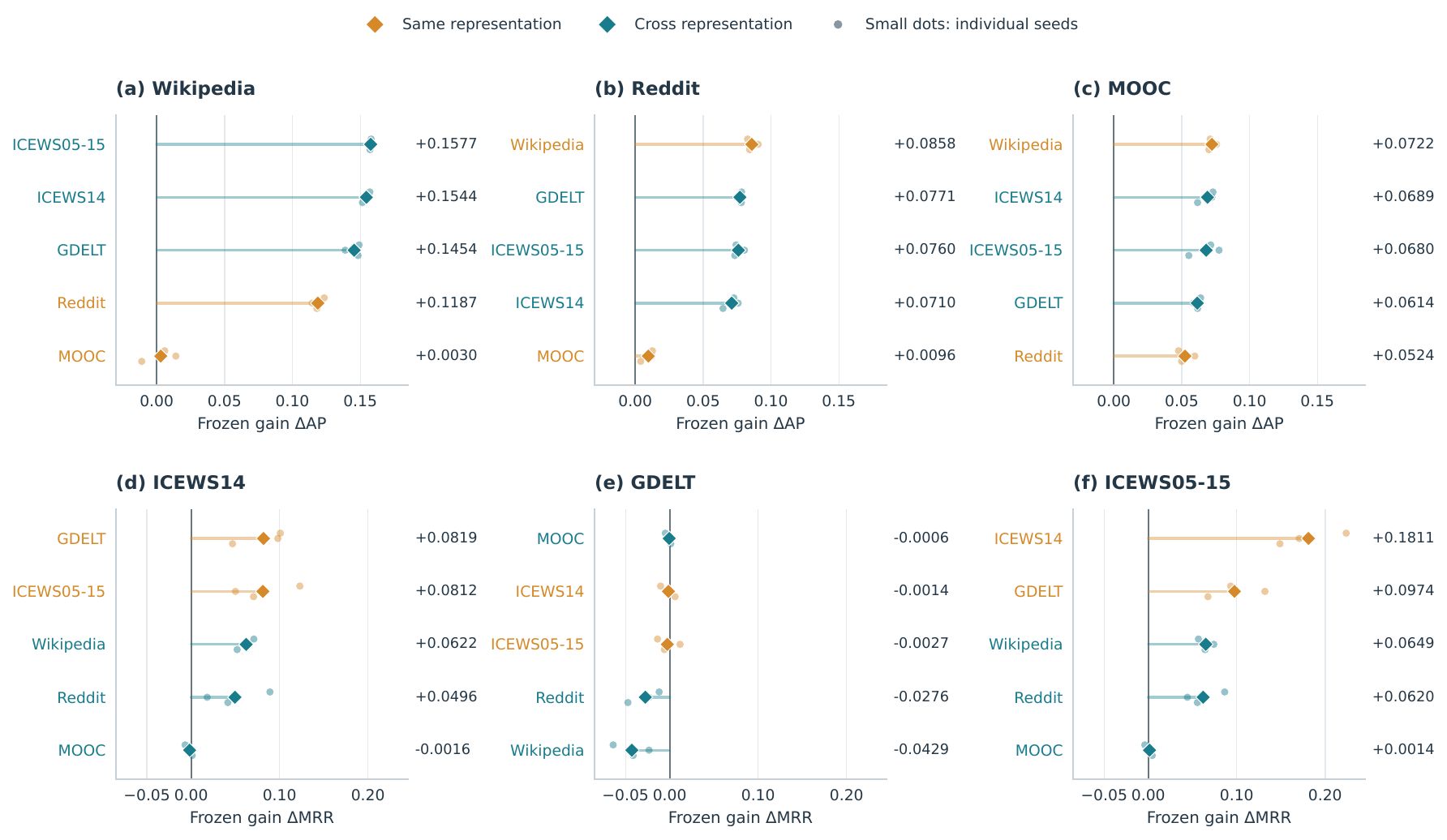}
\caption{\textbf{Source utility rankings across representation membership.} Sources are ordered by mean frozen gain (source-frozen $-$ random-frozen) for each target. Large diamonds denote three-seed means and small points individual seeds; orange denotes same-representation sources and teal cross-representation sources. For Wikipedia, all three cross-representation sources outperform both same-representation sources in mean gain. The top row reports AP for CTDG targets and the bottom row MRR for TKG targets, with common axis limits within each row.}
\label{fig:ordering}
\end{figure*}

Figure~\ref{fig:ordering} shows that cross-representation sources can provide higher utility than same-representation sources for the same target. For Wikipedia, all three TKG sources---ICEWS05-15 ($+0.1577$), ICEWS14 ($+0.1544$), and GDELT ($+0.1454$)---outperform both CTDG sources, Reddit ($+0.1187$) and MOOC ($+0.0030$), in mean gain. For Reddit, Wikipedia provides the highest gain, but all three TKG sources outperform the same-representation source MOOC. For MOOC, the three TKG sources rank after Wikipedia and before Reddit. Thus, for all three CTDG targets, at least one cross-representation source outperforms a same-representation source; for Wikipedia, every cross-representation source outperforms every same-representation source.

Restricting candidate pretraining sources by representation membership can exclude more useful learned knowledge. Restricting Wikipedia's candidates to CTDG sources would exclude all three highest observed transfer gains. This result shows that representation identity alone is insufficient for organizing sources of reusable knowledge.

Source orderings vary by target. Same-representation sources dominate for ICEWS14, where GDELT and ICEWS05-15 provide the two highest gains, and for ICEWS05-15, where ICEWS14 and GDELT do so. For GDELT, all sources yield nonpositive mean gains regardless of representation. Representation groupings and source utility rankings can therefore either align or cut across one another.

In summary, temporal graph representation can be associated with source utility but does not fully determine its ordering. Restricting foundation-model pretraining sources to representation-specific silos in advance can miss useful knowledge outside those silos. These findings motivate studying which relational dynamics are reusable between sources and targets as a principle for organizing learning resources.

\subsection{Unaligned Source-Learner Fusion}
\label{subsec:fusion}
We have established that predictive structure learned from different temporal graph representations can be reused across representation boundaries. We further test \textbf{whether predictive knowledge independently learned from different sources can be used jointly for a target query}. We pretrain history encoders separately on ICEWS05-15, a TKG of political events, and LastFM, a CTDG of music listening, and train target-specific interfaces using 5\% of Wikipedia training data. Source entities and relation semantics are not aligned; with each encoder frozen, we combine the two learners' standardized prediction scores. The TKG source and fusion weights are selected using validation data.

Table~\ref{tab:fusion} reports evaluation on a subsequent Wikipedia interval with the models and fusion weights fixed. The fused model achieves higher AP and AUC than either individual source in all three seeds. Mean AP increases from 0.7574 for LastFM alone to 0.7882, and AUC from 0.7859 to 0.8177. In particular, ICEWS05-15 provides additional predictive utility when combined with the LastFM learner despite its standalone AP of 0.6778.

\begin{table}[t]
\caption{Source-learner fusion from ICEWS05-15 and LastFM to Wikipedia (CTDG, 5\% training data). Historical-negative AP and AUC on a subsequent interval, reported as means $\pm$ sample standard deviations over seeds 7, 17, and 37.}
\label{tab:fusion}
\centering\scriptsize\setlength{\tabcolsep}{4pt}
\begin{tabular}{@{}lrr@{}}
\toprule
Source learner(s) & AP & AUC \\
\midrule
ICEWS05-15 (TKG) & $0.6778 \pm 0.0045$ & $0.6695 \pm 0.0130$ \\
LastFM (CTDG) & $0.7574 \pm 0.0164$ & $0.7859 \pm 0.0250$ \\
\textbf{ICEWS05-15 + LastFM} & $\boldsymbol{0.7882 \pm 0.0162}$ & $\boldsymbol{0.8177 \pm 0.0092}$ \\
\bottomrule
\end{tabular}
\end{table}

These results show that \textbf{predictive learners trained on relational histories from different domains and representations can be used jointly for a target query without aligning source data or entity and relation semantics}. Fusion combines the prediction scores produced by the transferred learners, rather than integrating the original graphs or their internal representation spaces.

\subsection{Practical Reuse Without Entity Alignment}
\label{subsec:practical_reuse}
We connect the reusability established above to utility in recommendation and security detection. Both PoCs fine-tune a pretrained history encoder on a target task without aligning source and target entities.

\subsubsection{Unaligned TKG Pretraining for Temporal Recommendation}
\label{subsec:recommendation}
Can semantically distant data, such as political events and music listening, be connected as learning resources? We fine-tune an ICEWS14-pretrained history encoder for next-item recommendation on LastFM. Source and target share neither entity and relation vocabularies nor item--entity alignment. We compare against the same randomly initialized model at 5\% and 20\% of target training data, evaluating all 1,000 items for 256 held-out test queries.

In Table~\ref{tab:recommendation}, Recall@10 improves in every seed at both training fractions. At 20\%, mean Recall@10 increases by 19.53 percentage points, and NDCG@10 also increases. At 5\%, NDCG@10 decreases, indicating different changes in hit rate and top-rank quality.

\begin{table}[t]
\caption{ICEWS14 $\rightarrow$ LastFM full fine-tuning. Each score cell reports Recall@10 / NDCG@10 as mean $\pm$ sample standard deviation. $\Delta$ is the within-seed difference between pretrained fine-tuning and target-only training, using seeds 7, 17, and 37.}
\label{tab:recommendation}
\centering\scriptsize
\setlength{\tabcolsep}{1.5pt}
\begin{tabular}{@{}lrrr@{}}
\toprule
Target train & Target-only & ICEWS14 fine-tuning & Paired $\Delta$ \\
\midrule
5\% · 1,147 & \makecell[r]{$0.2500\pm0.0000$ /\\$0.1981\pm0.0021$} & \makecell[r]{$0.3112\pm0.0158$ /\\$0.1693\pm0.0046$} & \makecell[r]{$+0.0612\pm0.0158$ /\\$-0.0287\pm0.0033$} \\
\addlinespace
20\% · 4,588 & \makecell[r]{$0.2708\pm0.0303$ /\\$0.2128\pm0.0291$} & \makecell[r]{$0.4661\pm0.0316$ /\\$0.2633\pm0.0391$} & \makecell[r]{$+0.1953\pm0.0517$ /\\$+0.0505\pm0.0609$} \\
\bottomrule
\end{tabular}
\end{table}

The improvements occur for repeat listening; the pretrained model's Recall@10 for previously unheard items is zero at both training fractions. \suppsec{sec:appendix_recommendation} provides results by recommendation situation and seed, along with the personal-recency heuristic, which achieves higher overall scores. This PoC shows that an external TKG without a semantic connection to the target can serve as a pretraining resource that improves the hit rate of the same recommendation model.

\subsubsection{External Threat Intelligence Before Integration}
\label{subsec:security}
This PoC connects relational dynamics reuse to an industrial question: can an external temporal source without a direct connection to the target system serve as a learning resource for internal attack detection without semantic alignment?

\begin{figure*}[t]
\centering
\includegraphics[width=\linewidth]{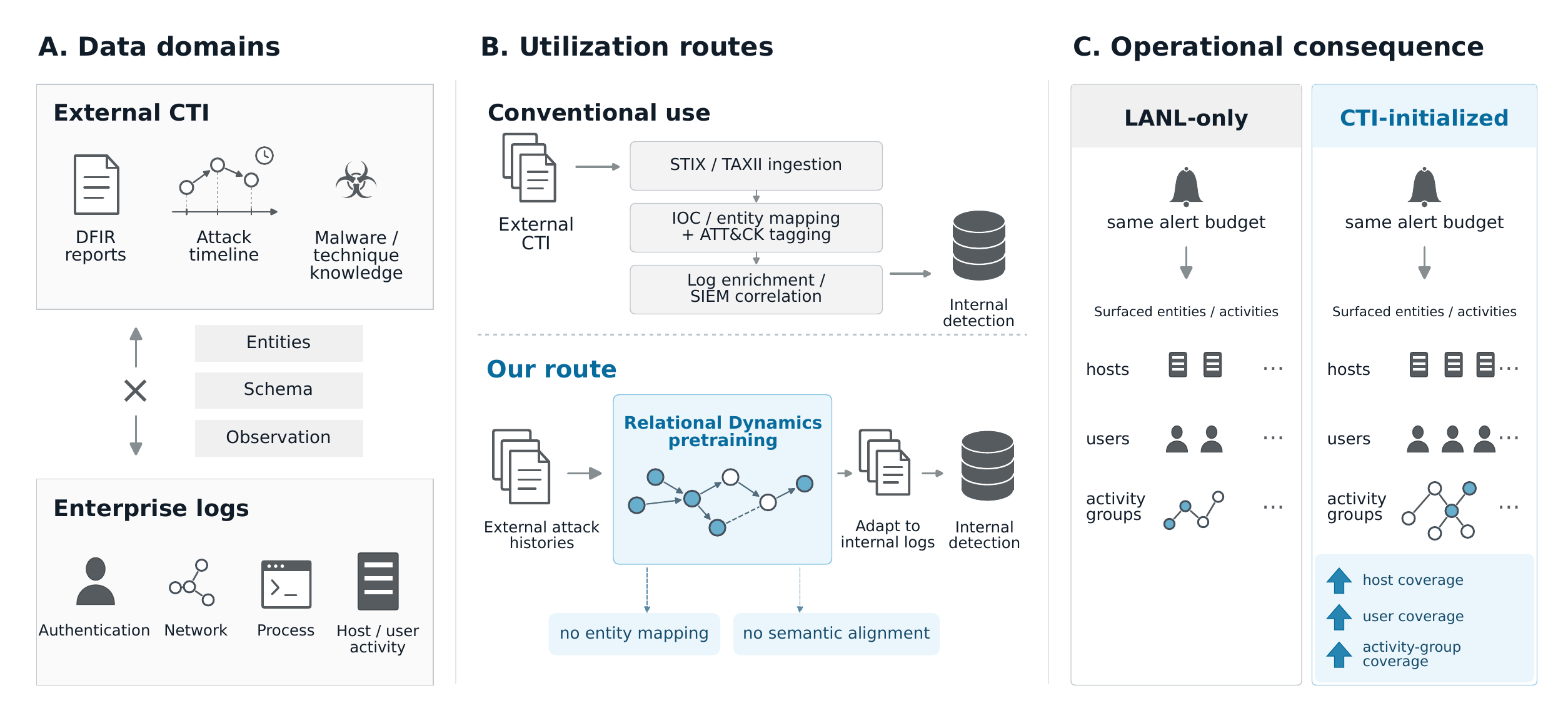}
\caption{Conceptual routes for using external CTI before integration. (a) Distinct external CTI and internal logs. (b) Learning and reusing relational dynamics without direct alignment. (c) Broader discovery of attack-related entities and activities under the same alert budget.}
\label{fig:security}
\end{figure*}

Figure~\ref{fig:security}(a) contrasts external threat intelligence with internal security logs. Malware, attack techniques, and intrusion procedures accumulate externally, while authentication, network, and process activities are recorded internally; the entities and schemas of the two sources do not directly correspond. Figure~\ref{fig:security}(b) shows two routes for using them. For example, Microsoft Sentinel supports STIX/TAXII-based CTI ingestion, management of ATT\&CK-related attack patterns and indicator enrichment, and alert generation by matching internal events against IOCs~\cite{microsoft2026sentinelti}. The conventional path in Figure~\ref{fig:security}(b) summarizes these integration functions. In addition to connecting the sources through indicator matching, technique tagging, and log enrichment, this PoC examines a route that pretrains relational dynamics on external attack histories and adapts them to internal logs. The key is to use external CTI as a learning resource before integration is complete.

Figure~\ref{fig:security}(c) represents this utility as discovery coverage under the same alert budget. Repeatedly detecting similar attack events and uncovering attack activity involving different users and hosts provide different information for investigation. We therefore evaluate attack-event counts alongside the coverage of unique hosts, users, and activity groups discovered within a limited number of alerts.

The source comprises 371 temporal-relational events directly extracted from 33 public intrusion reports published by The DFIR Report~\cite{dfirreport}, obtained through report links from ChronoCTI~\cite{rahman2025chronocti}. The target task is red-team event detection in LANL authentication logs. Only the encoder learned from report-specific strict-past histories is transferred; no entity mapping or common relation vocabulary is constructed. The LANL-only and CTI-initialized models undergo full fine-tuning for 10 epochs with identical target-private initialization and data. Source training duration is selected by report-level validation BCE, and the target checkpoint is selected by validation Recall@1\% FPR.

In Table~\ref{tab:security}, host, user, and activity-group coverage improves in all three seeds with 20\% training data under the same top-2500 alert budget. Mean host coverage increases from $6.67$ to $8.67$, with a paired increase of $+2.00\pm1.00$. More diverse attack-related entities and activities are uncovered even when fewer attack events are detected.

\begin{table}[t]
\caption{Comparison with 20\% LANL training data (29,847 events) and top-2500 alerts. Cells show LANL-only $\rightarrow$ CTI, with means in the final row. Hosts are deduplicated across both endpoints of detected red-team events. Activity groups are separated when the gap between attack events for the same user exceeds one hour.}
\label{tab:security}
\centering\scriptsize\setlength{\tabcolsep}{2pt}
\begin{tabular}{@{}lrrrr@{}}
\toprule
Seed & Hosts & Users & \makecell{Activity\\groups} & \makecell{Attack\\events} \\
\midrule
7 & $6\rightarrow8$ & $6\rightarrow7$ & $6\rightarrow7$ & $10\rightarrow7$ \\
17 & $6\rightarrow9$ & $5\rightarrow8$ & $6\rightarrow8$ & $7\rightarrow8$ \\
37 & $8\rightarrow9$ & $4\rightarrow8$ & $6\rightarrow8$ & $15\rightarrow10$ \\
\midrule
Mean & $6.67\rightarrow8.67$ & $5.00\rightarrow7.67$ & $6.00\rightarrow7.67$ & $10.67\rightarrow8.33$ \\
\bottomrule
\end{tabular}
\end{table}

\suppsec{sec:appendix_security} provides complete three-seed results across training fractions and alert budgets, along with the evaluation settings. At 5\%, mean host coverage increases from $2.00$ to $2.33$ for top-500 and top-1,000 alerts. At 100\%, however, mean event and host coverage is lower than the control at every evaluated alert budget. Alert counts refer to retrospective ranking of a test set with sampled background events, not actual daily operational alert volumes or confirmed intrusion incident counts.

These results show that reusing an encoder pretrained on an external temporal source without direct alignment can broaden discovery coverage in internal detection under some conditions with limited target data and alert budgets. This PoC alone cannot uniquely attribute the observed gains to relational dynamics. However, source and target use no entity, relation-vocabulary, or attack-technique alignment, and the preceding experiments established that cross-representation utility varies with source relational organization. The results are therefore consistent with the interpretation that relational dynamics learned from external temporal sources can contribute to this utility.

\section{Discussion and Limitation}
\label{sec:discussion}

Temporal graph representations are interfaces through which Relational Dynamics are observed and queried; they do not necessarily define the boundaries of what can be learned. The bidirectional reuse in Section~\ref{subsec:existence} and generalization across domains and training fractions in Section~\ref{subsec:scope} show that predictive structure learned from CTDGs and TKGs is not confined to its representation. Source utility rankings in the complete transfer matrix in Section~\ref{subsec:utility} also cross representation boundaries. These findings distinguish the divisions of a field that has treated CTDGs and TKGs separately from the boundaries of what its models learn.

The functional characterization further shows that reusable structure does not reduce to one inventory shared by every source and target. The original event sequence contributes consistently for ICEWS14 \(\rightarrow\) ICEWS05-15, whereas reorganizing event sequence, temporal placement, or relation direction provides higher utility for ICEWS05-15 \(\rightarrow\) Wikipedia. Learned and Reusable Relational Dynamics are therefore distinct: reuse depends not only on what is learned from a source, but also on the target and query to which it is applied.

Temporal graph foundation models should distinguish generalization across domains within one representation from learning Relational Dynamics that remain useful when the representation changes. Our results provide grounds for treating the latter as an independent learning objective. A future learner can retain representation-specific interfaces, accumulate Relational Dynamics from diverse sources, and select the part useful for a target condition.

Section~\ref{subsec:fusion} extends the reuse of relational dynamics to the combination of independently trained predictive learners from heterogeneous temporal sources. Predictive knowledge from different sources can thus be used jointly for a target query without first integrating data formats or entity and relation semantics. The experiment demonstrates utility at the score level; it does not identify a unique advantage of heterogeneous sources or a decomposition into specific dynamics mechanisms.

\subsection{Limitation}
\label{subsec:limitation}
Although this study poses relational dynamics reuse across temporal graph representations as a general research problem, its empirical analysis focuses on CTDGs and TKGs, which represent different observation and prediction interfaces in temporal graph learning. The results should therefore be interpreted as showing that representation differences are not absolute boundaries of learnable predictive structure between CTDGs and TKGs, rather than as claiming that all temporal graph representations share identical dynamics. Extending this perspective to other temporal graph representations remains future work.

\section{Conclusion}
\label{sec:conclusion}
This study demonstrates that aspects of Relational Dynamics can be learned and reused across temporal graph representations. Bidirectional frozen transfer between CTDGs and TKGs shows that learned predictive structure remains useful after changes in entity vocabulary and prediction interface. Controlled reorganizations of source histories show that the contributing organization varies across source--target conditions and that Learned Relational Dynamics are not reused in their entirety. Reuse across domains and training fractions and utility orderings across 30 source--target directions show that representation identity does not sufficiently explain the boundaries of learning and reuse.

These results extend the unit of study in temporal graph learning from representation-specific tasks to the relational dynamics reusable beneath them. Combining independently trained TKG and CTDG learners further shows that predictive knowledge can be used jointly on a third target without aligning the original sources' entities or relation semantics. The recommendation and security detection PoCs show that such reuse can extend to using temporal sources without direct alignment as new learning resources. The scope of a temporal graph foundation may be determined not only by which representations it supports, but also by which relational dynamics it can learn and reuse across different observations and queries.

\noindent\textbf{Acknowledgements.} The authors have no acknowledgements to declare.

\par\smallskip
\noindent\textbf{Funding.} This research received no specific grant from any funding
agency in the public, commercial, or not-for-profit sectors.

\par\smallskip
\noindent\textbf{Data availability.} The core experiments use seven public datasets:
Wikipedia, Reddit, MOOC, and LastFM from JODIE; ICEWS14 and ICEWS05-15 from
TLogic; and GDELT from CyGNet. \supptab{tab:data_sources} lists their sources.
The security PoC uses public LANL authentication and red-team data and timelines
extracted from public DFIR reports~\cite{dfirreport} located through
ChronoCTI~\cite{rahman2025chronocti}. Original reports remain with their
publishers, and the derived annotations are not included in the code release.
Retrieval, annotation, preprocessing, split, and evaluation instructions will be
provided with the experiment code.

\par\smallskip
\noindent\textbf{Code availability.} The implementation and experiment configurations will be made available at
\url{https://github.com/SnowyPainter/rdynamics-public}.

\appendices
\setcounter{table}{0}
\setcounter{figure}{0}
\setcounter{equation}{0}
\renewcommand{\thetable}{S\arabic{table}}
\renewcommand{\thefigure}{S\arabic{figure}}
\renewcommand{\theequation}{S\arabic{equation}}
\renewcommand{\theHtable}{S\arabic{table}}
\renewcommand{\theHfigure}{S\arabic{figure}}
\renewcommand{\theHequation}{S\arabic{equation}}
\section{Experimental Details}
\label{sec:appendix_details}

\paragraph{Training fractions and history.}
The 32,768-event splits preserve timestamp boundaries. The 5/20/100\% target training conditions retain the most recent training events, including all events at the boundary timestamp, so the actual fractions can be slightly larger. Excluded training events are removed from both training and evaluation histories. Sources use 100\% of their training data. Validation/test periods, query sets, and entity vocabularies remain identical across training fractions. During rolling evaluation, validation queries may use earlier validation observations, and test queries may use earlier validation and test observations; simultaneous and future events are excluded. Entity vocabularies are independent across datasets.

\paragraph{Candidates and filtering in the 32,768-event experiments.}
CTDG training draws one negative per positive uniformly with replacement from the destinations in the retained training data. Validation and test also sample one destination but use the destination set of the entire dataset; true or observed edges are not explicitly excluded. TKG training samples one entity uniformly per query, excluding training answers at the same timestamp. Validation uses the true answer and 100 distinct negatives; test uses all canonical entities. Validation/test filtering excludes other true answers with the same timestamp, relation, and fixed endpoint. Answer lists are constructed from all splits within the 32,768-event subset, including excluded training facts, but are separate from model input histories. We report optimistic ranks and also store average-tie ranks. CTDG evaluation RNG seeds are seed $+1000$ for validation and seed $+2000$ for test; the TKG validation seed is 123456.

\paragraph{Training budget.}
The 10 epochs are a fixed budget chosen for computational cost, not early stopping. The formal transfer experiments select the validation-best checkpoint. Equal event and epoch budgets do not imply equal numbers of optimizer updates across representations. The GraphMixer and TGAT controls are random-frozen instances of the corresponding backbone.

\paragraph{Bidirectional DyGFormer transfer settings.}
Sections~\ref{subsec:existence} and~\ref{subsec:organization} train on the full target training set for 10 epochs. ICEWS14 has 63,685/13,823/13,222 training/validation/test facts, and Wikipedia has 70,000/15,000/15,000 events. Table~\ref{tab:appendix_protocol} compares checkpoint selection and evaluation scope across experiments.

\begin{table*}[t]
\caption{Checkpoint selection and evaluation scope for bidirectional DyGFormer experiments.}
\label{tab:appendix_protocol}
\small\renewcommand{\arraystretch}{1.2}
\begin{tabularx}{\linewidth}{@{}lXXX@{}}
\toprule
Direction & Target checkpoint selection & Existence evaluation & Source organization evaluation \\
\midrule
Wiki $\rightarrow$ ICEWS14 & 100-negative MRR on 256 bidirectional queries from 128 fixed validation facts & Full test: 26,444 queries $\times$ 7,128 entities & Same 256 validation queries used for selection $\times$ 7,128 entities \\
ICEWS14 $\rightarrow$ Wiki & AP on all 15,000 validation events & AP and AUC on all 15,000 test events & AP and AUC on all 15,000 test events \\
\bottomrule
\end{tabularx}
\end{table*}

The existence experiment evaluates previously validation-selected checkpoints on the full test set without retraining or reselection. ICEWS14 test filtering excludes other true answers with the same timestamp, relation, and fixed endpoint across the original training, validation, and test splits; this answer list is separate from the strict-past input history. We report optimistic ranks and also store average-tie ranks. The source-organization experiment on ICEWS14 uses timestamp filtering based only on validation answers and average-tie ranks. Wikipedia uses one random destination per positive as a negative.

The reverse-direction source is an ICEWS14 checkpoint without prior Wiki pretraining. The intact, time-shuffled, and partner-shuffled sources in the source-organization experiment are trained from identical initialization for 10 epochs, and their final checkpoints are transferred. Time shuffle preserves directed-pair and relation frequencies and the timestamp distribution; partner shuffle preserves node activity within time bins.

\paragraph{Partial partner-corruption control.}
Within each timestamp--relation group in ICEWS14 training data, events are selected with the specified probability, and their destinations are permuted among the selected events. Source initialization, optimizer, and training budget remain identical, and the final epoch-10 checkpoint is used. All source models are evaluated with the original ICEWS14 history and identical queries and negative candidates. Strength selection uses 2,048 bidirectional queries from 1,024 validation facts that do not overlap the earlier 256-query diagnostic set (fact-sampling seed 20261005, 100-negative sampling seed 654321, average-tie ranks). Because this set is used to select strength, it is not treated as an independent validation set.

The relative difference is computed as \(\lvert M_S^{\mathrm{partial}}-M_S^{\mathrm{intact}}\rvert/M_S^{\mathrm{intact}}\). We select the strongest common level whose difference is within 3\% for all three seeds. Table~\ref{tab:appendix_strength} yields a selected level of 50\%.

\begin{table*}[t]
\caption{Selection of partial corruption strength based on source performance.}
\label{tab:appendix_strength}
\centering\small
\begin{tabular}{@{}llll@{}}
\toprule
Nominal strength & Relative source MRR difference: seeds 7 / 17 / 37 & All seeds within tolerance & Selected \\
\midrule
25\% & 0.725\% / 0.584\% / 0.832\% & Yes & --- \\
50\% & 0.503\% / 0.252\% / 0.946\% & Yes & Yes \\
75\% & 2.817\% / 4.774\% / 3.610\% & No & --- \\
\bottomrule
\end{tabular}
\end{table*}

This grid was designed as a follow-up after observing the target pilot at 25\% corruption with seed 7. We then fixed the grid selection rule, selected 50\% using source scores alone, and evaluated target results for that strength only after selection. We reuse the intact target results under the same conditions and apply identical target-private initialization, training budget, validation selection, and full-test evaluation to the corrupted condition. Corrupted event sets are not guaranteed to be nested across strengths, and results at one selected strength are not interpreted as a target dose--response.

\section{Full Results Across Domains and Data Regimes}
\label{sec:appendix_results}
This appendix provides the complete results for Section~\ref{subsec:scope}. Each value is the mean $\pm$ sample standard deviation across seeds 7, 17, and 37; paired gains are computed between matched conditions within each seed.

\paragraph{Frozen transfer.}
Table~\ref{tab:appendix_frozen} provides the full scores for each condition. R denotes random-frozen, S denotes source-frozen, and parentheses show paired gains (S $-$ R).

\begin{table*}[t]
\caption{Full frozen-transfer scores and paired gains.}
\label{tab:appendix_frozen}
\centering\small\setlength{\tabcolsep}{5pt}\renewcommand{\arraystretch}{1.15}
\begin{tabular}{@{}llccc@{}}
\toprule
Source $\rightarrow$ Target & Metric & 5\% train & 20\% train & 100\% train \\
\midrule
Wikipedia $\rightarrow$ ICEWS14 & MRR &
\makecell[l]{R $0.0627\pm0.0093$\\S $0.1121\pm0.0079$\\$\mathbf{(+0.0494\pm0.0130)}$} &
\makecell[l]{R $0.0709\pm0.0054$\\S $0.1384\pm0.0059$\\$\mathbf{(+0.0675\pm0.0096)}$} &
\makecell[l]{R $0.0594\pm0.0009$\\S $0.1216\pm0.0093$\\$\mathbf{(+0.0622\pm0.0095)}$} \\
\addlinespace
Reddit $\rightarrow$ ICEWS14 & MRR &
\makecell[l]{R $0.0627\pm0.0093$\\S $0.1187\pm0.0152$\\$\mathbf{(+0.0559\pm0.0064)}$} &
\makecell[l]{R $0.0709\pm0.0054$\\S $0.1180\pm0.0247$\\$\mathbf{(+0.0471\pm0.0259)}$} &
\makecell[l]{R $0.0594\pm0.0009$\\S $0.1090\pm0.0368$\\$\mathbf{(+0.0496\pm0.0362)}$} \\
\addlinespace
MOOC $\rightarrow$ GDELT & MRR &
\makecell[l]{R $0.1123\pm0.0124$\\S $0.0714\pm0.0129$\\$\mathbf{(-0.0409\pm0.0128)}$} &
\makecell[l]{R $0.0836\pm0.0304$\\S $0.0423\pm0.0100$\\$\mathbf{(-0.0413\pm0.0380)}$} &
\makecell[l]{R $0.1056\pm0.0175$\\S $0.1050\pm0.0142$\\$\mathbf{(-0.0006\pm0.0038)}$} \\
\addlinespace
ICEWS14 $\rightarrow$ Wikipedia & AP &
\makecell[l]{R $0.7314\pm0.0154$\\S $0.9359\pm0.0025$\\$\mathbf{(+0.2046\pm0.0134)}$} &
\makecell[l]{R $0.7338\pm0.0076$\\S $0.9476\pm0.0060$\\$\mathbf{(+0.2138\pm0.0034)}$} &
\makecell[l]{R $0.8099\pm0.0028$\\S $0.9643\pm0.0051$\\$\mathbf{(+0.1544\pm0.0028)}$} \\
\addlinespace
GDELT $\rightarrow$ Reddit & AP &
\makecell[l]{R $0.8564\pm0.0067$\\S $0.9117\pm0.0050$\\$\mathbf{(+0.0553\pm0.0103)}$} &
\makecell[l]{R $0.8614\pm0.0062$\\S $0.9306\pm0.0027$\\$\mathbf{(+0.0691\pm0.0042)}$} &
\makecell[l]{R $0.8784\pm0.0013$\\S $0.9555\pm0.0033$\\$\mathbf{(+0.0771\pm0.0020)}$} \\
\addlinespace
Wikipedia $\rightarrow$ ICEWS05-15 & MRR &
\makecell[l]{R $0.0718\pm0.0012$\\S $0.1172\pm0.0091$\\$\mathbf{(+0.0454\pm0.0086)}$} &
\makecell[l]{R $0.0929\pm0.0039$\\S $0.1439\pm0.0154$\\$\mathbf{(+0.0510\pm0.0162)}$} &
\makecell[l]{R $0.0645\pm0.0013$\\S $0.1294\pm0.0079$\\$\mathbf{(+0.0649\pm0.0087)}$} \\
\addlinespace
ICEWS05-15 $\rightarrow$ MOOC & AP &
\makecell[l]{R $0.6094\pm0.0029$\\S $0.6548\pm0.0288$\\$\mathbf{(+0.0453\pm0.0308)}$} &
\makecell[l]{R $0.6308\pm0.0034$\\S $0.6701\pm0.0177$\\$\mathbf{(+0.0394\pm0.0145)}$} &
\makecell[l]{R $0.6175\pm0.0082$\\S $0.6855\pm0.0110$\\$\mathbf{(+0.0680\pm0.0115)}$} \\
\bottomrule
\end{tabular}
\end{table*}

\paragraph{Fine-tuning.}
Table~\ref{tab:appendix_finetune} provides the full scores for each condition. Source denotes source-finetune, and paired gains are source-finetune $-$ target-only.

\begin{table*}[t]
\caption{Full fine-tuning scores and paired gains.}
\label{tab:appendix_finetune}
\centering\small\setlength{\tabcolsep}{4pt}\renewcommand{\arraystretch}{1.1}
\begin{tabular}{@{}llrrrr@{}}
\toprule
Source $\rightarrow$ Target & Metric & Train & Target-only & Source & Paired $\Delta$ \\
\midrule
Wikipedia $\rightarrow$ ICEWS14 & MRR & 5\% & $0.1537\pm0.0330$ & $0.1348\pm0.0183$ & $-0.0189\pm0.0222$ \\
Wikipedia $\rightarrow$ ICEWS14 & MRR & 20\% & $0.1499\pm0.0290$ & $0.1779\pm0.0286$ & $+0.0280\pm0.0277$ \\
Wikipedia $\rightarrow$ ICEWS14 & MRR & 100\% & $0.1860\pm0.0361$ & $0.2025\pm0.0218$ & $+0.0165\pm0.0531$ \\
\addlinespace
Reddit $\rightarrow$ ICEWS14 & MRR & 5\% & $0.1537\pm0.0330$ & $0.1512\pm0.0347$ & $-0.0024\pm0.0057$ \\
Reddit $\rightarrow$ ICEWS14 & MRR & 20\% & $0.1499\pm0.0290$ & $0.1497\pm0.0198$ & $-0.0002\pm0.0159$ \\
Reddit $\rightarrow$ ICEWS14 & MRR & 100\% & $0.1860\pm0.0361$ & $0.2231\pm0.0377$ & $+0.0371\pm0.0308$ \\
\addlinespace
MOOC $\rightarrow$ GDELT & MRR & 5\% & $0.1178\pm0.0146$ & $0.0977\pm0.0267$ & $-0.0201\pm0.0137$ \\
MOOC $\rightarrow$ GDELT & MRR & 20\% & $0.0853\pm0.0240$ & $0.1083\pm0.0080$ & $+0.0230\pm0.0294$ \\
MOOC $\rightarrow$ GDELT & MRR & 100\% & $0.1238\pm0.0092$ & $0.1202\pm0.0084$ & $-0.0036\pm0.0098$ \\
\addlinespace
ICEWS14 $\rightarrow$ Wikipedia & AP & 5\% & $0.7483\pm0.0068$ & $0.9533\pm0.0032$ & $+0.2049\pm0.0040$ \\
ICEWS14 $\rightarrow$ Wikipedia & AP & 20\% & $0.9400\pm0.0191$ & $0.9707\pm0.0018$ & $+0.0307\pm0.0209$ \\
ICEWS14 $\rightarrow$ Wikipedia & AP & 100\% & $0.9836\pm0.0015$ & $0.9855\pm0.0018$ & $+0.0019\pm0.0015$ \\
\addlinespace
GDELT $\rightarrow$ Reddit & AP & 5\% & $0.8760\pm0.0049$ & $0.9260\pm0.0040$ & $+0.0500\pm0.0076$ \\
GDELT $\rightarrow$ Reddit & AP & 20\% & $0.8982\pm0.0018$ & $0.9425\pm0.0021$ & $+0.0443\pm0.0018$ \\
GDELT $\rightarrow$ Reddit & AP & 100\% & $0.9578\pm0.0020$ & $0.9673\pm0.0011$ & $+0.0095\pm0.0029$ \\
\addlinespace
Wikipedia $\rightarrow$ ICEWS05-15 & MRR & 5\% & $0.1347\pm0.0186$ & $0.1591\pm0.0192$ & $+0.0244\pm0.0337$ \\
Wikipedia $\rightarrow$ ICEWS05-15 & MRR & 20\% & $0.1527\pm0.0236$ & $0.2017\pm0.0177$ & $+0.0491\pm0.0134$ \\
Wikipedia $\rightarrow$ ICEWS05-15 & MRR & 100\% & $0.1868\pm0.0507$ & $0.2383\pm0.0395$ & $+0.0515\pm0.0482$ \\
\addlinespace
ICEWS05-15 $\rightarrow$ MOOC & AP & 5\% & $0.6363\pm0.0058$ & $0.6926\pm0.0088$ & $+0.0563\pm0.0139$ \\
ICEWS05-15 $\rightarrow$ MOOC & AP & 20\% & $0.6334\pm0.0001$ & $0.7120\pm0.0069$ & $+0.0786\pm0.0070$ \\
ICEWS05-15 $\rightarrow$ MOOC & AP & 100\% & $0.6351\pm0.0052$ & $0.7127\pm0.0064$ & $+0.0776\pm0.0101$ \\
\bottomrule
\end{tabular}
\end{table*}

\section{Cross-Backbone Transfer Results}
\label{sec:appendix_backbones}
Table~\ref{tab:backbones} provides the scores and paired gains underlying the GraphMixer and TGAT results in Figure~\ref{fig:backbones}. Scores are means across seeds 7, 17, and 37, and paired gains are the mean $\pm$ sample standard deviation of within-seed source-frozen $-$ random-frozen differences. DyGFormer scores appear in Table~\ref{tab:existence} in the main text.

\begin{table*}[t]
\caption{Bidirectional frozen transfer with GraphMixer and TGAT.}
\label{tab:backbones}
\centering\small\setlength{\tabcolsep}{4pt}
\begin{tabular}{@{}lllrrr@{}}
\toprule
Backbone & Direction & Test metric & Random-frozen & Source-frozen & Paired $\Delta\pm$ SD \\
\midrule
GraphMixer & Wiki $\rightarrow$ ICEWS14 & MRR & 0.0529 & 0.0431 & $-0.0099\pm0.0119$ \\
GraphMixer & ICEWS14 $\rightarrow$ Wiki & AP & 0.7822 & 0.8042 & $\mathbf{+0.0219\pm0.0163}$ \\
TGAT & Wiki $\rightarrow$ ICEWS14 & MRR & 0.0155 & 0.0442 & $\mathbf{+0.0287\pm0.0065}$ \\
TGAT & ICEWS14 $\rightarrow$ Wiki & AP & 0.7810 & 0.8956 & $\mathbf{+0.1146\pm0.0548}$ \\
\bottomrule
\end{tabular}
\end{table*}

\section{Recommendation PoC Details}
\label{sec:appendix_recommendation}
Table~\ref{tab:recommendation_seeds} provides per-seed results for Section~\ref{subsec:recommendation}. All conditions use identical validation/test queries and are evaluated with seeds 7, 17, and 37. Training fractions refer to supervision and initial training history, not reductions in the validation/test interactions observed during evaluation. An unheard item is one for which the corresponding user--item interaction is absent from the permitted history preceding the query (5\%: 61 queries; 20\%: 59 queries).

\begin{table*}[t]
\caption{Per-seed full fine-tuning results for the recommendation PoC. Each cell reports Recall@10 / NDCG@10.}
\label{tab:recommendation_seeds}
\centering\small
\begin{tabular}{@{}llrr@{}}
\toprule
Target train & Seed & Target-only & ICEWS14 fine-tuning \\
\midrule
5\% & 7 & 0.2500 / 0.2004 & 0.2930 / 0.1741 \\
5\% & 17 & 0.2500 / 0.1975 & 0.3203 / 0.1649 \\
5\% & 37 & 0.2500 / 0.1964 & 0.3203 / 0.1690 \\
20\% & 7 & 0.3047 / 0.2464 & 0.4609 / 0.2395 \\
20\% & 17 & 0.2617 / 0.1982 & 0.4375 / 0.2420 \\
20\% & 37 & 0.2461 / 0.1939 & 0.5000 / 0.3083 \\
\bottomrule
\end{tabular}
\end{table*}

Table~\ref{tab:recommendation_heuristics} reports heuristic scores on the same queries and candidate sets. Popularity ranks items by cumulative interaction count, personal frequency by each user's listening count per item, and personal recency by each user's most recent listening time per item. All use only retained training history and validation/test history preceding the query. Heuristics use the entire permitted past, whereas the model uses up to 63 recent events per endpoint. Tie handling is identical to that of the model, and the heuristics have no training seeds.

\begin{table*}[t]
\caption{Recommendation heuristics under identical evaluation conditions. Each cell reports Recall@10 / NDCG@10.}
\label{tab:recommendation_heuristics}
\centering\small
\begin{tabular}{@{}lrr@{}}
\toprule
Method & 5\% train history & 20\% train history \\
\midrule
Item popularity & 0.2461 / 0.1825 & 0.2344 / 0.1718 \\
Personal frequency & 0.4297 / 0.2829 & 0.4102 / 0.2625 \\
Personal recency & 0.5078 / 0.3464 & 0.5078 / 0.3464 \\
\bottomrule
\end{tabular}
\end{table*}

Table~\ref{tab:recommendation_groups} reports reuse gains by recommendation situation. Repeat listening denotes a query whose user--item interaction occurs in the permitted history preceding that query.

\begin{table*}[t]
\caption{Means $\pm$ sample standard deviations and paired gains by recommendation situation. Each score cell reports Recall@10 / NDCG@10; parentheses indicate query counts.}
\label{tab:recommendation_groups}
\centering\small
\setlength{\tabcolsep}{3pt}
\begin{tabular}{@{}llrrr@{}}
\toprule
Target train & Situation (queries) & Target-only & ICEWS14 fine-tuning & Paired $\Delta$ \\
\midrule
5\% & Repeat listening (195) & \makecell[r]{$0.3231\pm0.0000$ /\\$0.2574\pm0.0023$} & \makecell[r]{$0.4085\pm0.0207$ /\\$0.2223\pm0.0060$} & \makecell[r]{$+0.0855\pm0.0207$ /\\$-0.0351\pm0.0050$} \\
\addlinespace
5\% & Unheard items (61) & \makecell[r]{$0.0164\pm0.0000$ /\\$0.0083\pm0.0020$} & \makecell[r]{$0.0000\pm0.0000$ /\\$0.0000\pm0.0000$} & \makecell[r]{$-0.0164\pm0.0000$ /\\$-0.0083\pm0.0020$} \\
\addlinespace
20\% & Repeat listening (197) & \makecell[r]{$0.3469\pm0.0394$ /\\$0.2745\pm0.0384$} & \makecell[r]{$0.6058\pm0.0410$ /\\$0.3421\pm0.0508$} & \makecell[r]{$+0.2589\pm0.0672$ /\\$+0.0676\pm0.0795$} \\
\addlinespace
20\% & Unheard items (59) & \makecell[r]{$0.0169\pm0.0000$ /\\$0.0069\pm0.0018$} & \makecell[r]{$0.0000\pm0.0000$ /\\$0.0000\pm0.0000$} & \makecell[r]{$-0.0169\pm0.0000$ /\\$-0.0069\pm0.0018$} \\
\bottomrule
\end{tabular}
\end{table*}

\section{Security PoC Details}
\label{sec:appendix_security}
Table~\ref{tab:security_regimes} provides the complete results across training fractions and alert budgets for Section~\ref{subsec:security}. Target training subsets are nested samples drawn using a fixed, label-blind permutation; unselected training events are excluded from both supervision and history. Validation and test sets remain unchanged, and only observed strict-past events are added to the history. Source training duration is selected by BCE over epochs 0--100 using 26 training reports and seven validation reports, followed by retraining on all 33 reports.

The first 30 days of LANL are split into training $[0,9)$, validation $[9,14)$, and test $[14,30)$ days. Events matching the public red-team records are retained, and the remainder are deterministically sampled at 1/1024. The test set contains 263,317 events, including 104 red-team events covering 71 hosts, 51 users, and 66 activity groups. Host counts denote the endpoints of attack events, not confirmed compromise of each host; activity groups are likewise not actual incident IDs. Top-K denotes retrospective ranking over the entire sampled test set.

\begin{table*}[t]
\caption{Means across seeds 7, 17, and 37 by LANL training fraction and alert budget. Each cell shows LANL-only $\rightarrow$ CTI. Training fractions of 5\%, 20\%, and 100\% contain 7,461, 29,847, and 149,237 events, respectively.}
\label{tab:security_regimes}
\centering\small
\setlength{\tabcolsep}{4pt}
\begin{tabular}{@{}lrrrrr@{}}
\toprule
Target train & Alerts K & Hosts & Users & Activity groups & Attack events \\
\midrule
5\% & 100 & $1.33\rightarrow1.33$ & $0.67\rightarrow0.67$ & $0.67\rightarrow0.67$ & $0.67\rightarrow0.67$ \\
5\% & 500 & $2.00\rightarrow2.33$ & $1.00\rightarrow1.33$ & $1.00\rightarrow1.33$ & $1.00\rightarrow1.33$ \\
5\% & 1,000 & $2.00\rightarrow2.33$ & $1.00\rightarrow1.33$ & $1.00\rightarrow1.33$ & $1.00\rightarrow1.33$ \\
5\% & 2,500 & $6.33\rightarrow5.67$ & $6.33\rightarrow4.67$ & $6.33\rightarrow4.67$ & $6.33\rightarrow5.00$ \\
\addlinespace
20\% & 100 & $0.00\rightarrow0.67$ & $0.00\rightarrow0.33$ & $0.00\rightarrow0.33$ & $0.00\rightarrow0.33$ \\
20\% & 500 & $3.00\rightarrow2.33$ & $2.00\rightarrow1.67$ & $2.33\rightarrow1.67$ & $3.67\rightarrow1.67$ \\
20\% & 1,000 & $3.33\rightarrow3.33$ & $2.67\rightarrow2.67$ & $3.00\rightarrow2.67$ & $5.67\rightarrow2.67$ \\
20\% & 2,500 & $6.67\rightarrow8.67$ & $5.00\rightarrow7.67$ & $6.00\rightarrow7.67$ & $10.67\rightarrow8.33$ \\
\addlinespace
100\% & 100 & $16.33\rightarrow15.67$ & $9.00\rightarrow8.67$ & $12.00\rightarrow11.00$ & $20.67\rightarrow18.00$ \\
100\% & 500 & $28.00\rightarrow27.00$ & $18.00\rightarrow15.00$ & $22.33\rightarrow21.00$ & $35.33\rightarrow34.67$ \\
100\% & 1,000 & $34.67\rightarrow33.33$ & $23.00\rightarrow19.00$ & $28.67\rightarrow26.67$ & $43.33\rightarrow42.00$ \\
100\% & 2,500 & $42.67\rightarrow39.00$ & $29.33\rightarrow23.33$ & $36.67\rightarrow33.00$ & $55.67\rightarrow52.67$ \\
\bottomrule
\end{tabular}
\end{table*}

\section{Same-Dataset Representation Diagnostic}
\label{sec:appendix_same_dataset}
Table~\ref{tab:same_dataset} reports the three-seed results for the diagnostic in Section~\ref{subsec:existence}.

The same-dataset diagnostic uses a shared MLP scorer over history frequency and recency statistics and adapts the entire scorer, separately from frozen transfer. Both source conditions use matched initialization, optimizer, negative sampling, and training budgets, with 10 pretraining epochs on the same 50,080 earlier events. The relation-hidden condition disables relation-specific statistics while retaining endpoints, timestamps, and event multiplicity. All conditions then adapt for 10 epochs on the same 735 subsequent events (approximately 1\% of training events), with the source period supplied as common input history only. Source, adaptation, validation, and test periods are chronologically separated; evaluation uses bidirectional timestamp-filtered MRR on the full test set.

\begin{table*}[t]
\caption{Same-dataset pretraining with relations retained or hidden in ICEWS14. Values are means $\pm$ standard deviations over three seeds; gains are relative to scratch.}
\label{tab:same_dataset}
\centering\small
\begin{tabular}{@{}lrr@{}}
\toprule
Pretraining condition & Test MRR & Gain \\
\midrule
No pretraining & $0.3551 \pm 0.0171$ & --- \\
TKG: relations retained & $0.4510 \pm 0.0010$ & $+0.0959$ \\
CTDG-style: relations hidden & $0.4297 \pm 0.0039$ & $+0.0746$ \\
\bottomrule
\end{tabular}
\end{table*}

\section{Functional Profiles of Reusable Relational Dynamics}
\label{sec:appendix_functional_profiles}
This appendix reports candidate regularities in the data, the construction and distributional effects of controlled variants, comparisons on the original source task, and the complete source--target results. Table~\ref{tab:functional_profiles} in the main text summarizes conditions with consistent directions across seeds and clear effect sizes.

\paragraph{Candidate regularities in the data.}
The heuristics in Table~\ref{tab:functional_heuristics} use only observations preceding each query and show that different predictive regularities are strong in different datasets. They describe candidate structure in source histories; the following controlled variants distinguish structure learned and reused by the model.

\begin{table*}[t]
\caption{Strict-past heuristic AUC by dataset. Parentheses for pair renewal show the fraction of applicable queries.}
\label{tab:functional_heuristics}
\centering\scriptsize
\resizebox{\linewidth}{!}{%
\begin{tabular}{@{}lrrrrrrrrr@{}}
\toprule
Dataset & Pair frequency & Pair recency & Activity & Recency & Latest partner & 1-step transition & Pair renewal & Reciprocity & Neighbor overlap \\
\midrule
Wikipedia & .631 & .771 & .493 & .826 & .805 & .775 & .781 (60.8\%) & -- & -- \\
Reddit & .423 & .370 & .528 & .504 & .475 & .519 & .591 (25.6\%) & -- & -- \\
MOOC & .313 & .331 & .414 & .493 & .483 & .536 & .611 (14.8\%) & -- & -- \\
ICEWS14 & .389 & .390 & .528 & .541 & .462 & .518 & .699 (14.6\%) & .428 & .570 \\
GDELT & .195 & .175 & .464 & .474 & .366 & .491 & .539 (3.3\%) & .342 & .533 \\
ICEWS05-15 & .485 & .457 & .562 & .571 & .506 & .553 & .628 (21.1\%) & .498 & .573 \\
\bottomrule
\end{tabular}}
\end{table*}

\paragraph{Controlled variants and source competence.}
Table~\ref{tab:functional_controls} defines the history organization preserved and reorganized by each variant, and Table~\ref{tab:functional_audit} reports the resulting changes in pair and transition distributions. The original and variant encoders are evaluated on the same original source history, and only conditions whose relative difference in Eq.~\eqref{eq:source_gap} is at most 3\% for every seed are transferred. Thresholds of 1\%, 2\%, and 3\% retain 8, 14, and 15 conditions, respectively. The 3\% criterion was fixed before examining target results.

\begin{table*}[t]
\caption{Preserved statistics and reorganized history aspects for each controlled variant.}
\label{tab:functional_controls}
\centering\small
\begin{tabularx}{\linewidth}{@{}lXX@{}}
\toprule
Variant & Preserved & Reorganized \\
\midrule
Partner & Event count, relation frequencies, timestamp multiset, and binned destination frequencies & Directed partner pairs and successive transitions \\
Sequence & Directed typed-pair frequencies and source--relation time slots & Partner placement and order across slots \\
Time & Directed typed-pair frequencies and timestamp multiset & Assignment of complete relational events to times \\
Direction (TKG) & Undirected typed pairs, relation frequencies, and timestamps & Subject--object direction \\
\bottomrule
\end{tabularx}
\end{table*}

\begin{table*}[t]
\caption{Performance differences between controlled-variant and original encoders on the original source history.}
\label{tab:functional_competence}
\centering\scriptsize
\begin{tabular}{@{}llrrc@{}}
\toprule
Source & Variant & Mean gap & Maximum gap & 3\% criterion \\
\midrule
Wikipedia & Partner & 31.70\% & 34.67\% & Excluded \\
Wikipedia & Sequence & 0.01\% & 0.02\% & Included \\
Wikipedia & Time & 0.17\% & 0.36\% & Included \\
Reddit & Partner & 7.02\% & 7.65\% & Excluded \\
Reddit & Sequence & 0.19\% & 0.35\% & Included \\
Reddit & Time & 0.66\% & 0.99\% & Included \\
MOOC & Partner & 0.28\% & 0.39\% & Included \\
MOOC & Sequence & 0.19\% & 0.33\% & Included \\
MOOC & Time & 0.70\% & 1.29\% & Included \\
ICEWS14 & Partner & 14.62\% & 14.62\% & Excluded \\
ICEWS14 & Sequence & 0.65\% & 0.81\% & Included \\
ICEWS14 & Time & 0.79\% & 1.31\% & Included \\
ICEWS14 & Direction & 0.85\% & 1.21\% & Included \\
GDELT & Partner & 3.16\% & 3.16\% & Excluded \\
GDELT & Sequence & 1.32\% & 2.20\% & Included \\
GDELT & Time & 2.05\% & 3.41\% & Excluded \\
GDELT & Direction & 0.61\% & 1.05\% & Included \\
ICEWS05-15 & Partner & 11.15\% & 11.15\% & Excluded \\
ICEWS05-15 & Sequence & 0.83\% & 1.31\% & Included \\
ICEWS05-15 & Time & 0.50\% & 0.72\% & Included \\
ICEWS05-15 & Direction & 0.83\% & 1.25\% & Included \\
\bottomrule
\end{tabular}
\end{table*}

\begin{table*}[t]
\caption{Total variation (TV) between pair and transition count distributions before and after each controlled reorganization.}
\label{tab:functional_audit}
\centering\scriptsize
\begin{tabular}{@{}llrr@{}}
\toprule
Source & Variant & Pair TV & Transition TV \\
\midrule
Wikipedia & Partner & .971 & .998 \\
Wikipedia & Sequence & .000 & .161 \\
Wikipedia & Time & .000 & .165 \\
Reddit & Partner & .955 & .997 \\
Reddit & Sequence & .000 & .249 \\
Reddit & Time & .000 & .254 \\
MOOC & Partner & .493 & .923 \\
MOOC & Sequence & .000 & .759 \\
MOOC & Time & .000 & .757 \\
ICEWS14 & Partner & .820 & .956 \\
ICEWS14 & Sequence & .000 & .472 \\
ICEWS14 & Time & .000 & .465 \\
ICEWS14 & Direction & .737 & .788 \\
GDELT & Partner & .796 & .967 \\
GDELT & Sequence & .000 & .698 \\
GDELT & Time & .000 & .689 \\
GDELT & Direction & .795 & .933 \\
ICEWS05-15 & Partner & .782 & .941 \\
ICEWS05-15 & Sequence & .000 & .459 \\
ICEWS05-15 & Time & .000 & .455 \\
ICEWS05-15 & Direction & .687 & .739 \\
\bottomrule
\end{tabular}
\end{table*}

\paragraph{Complete target results.}
For each source, the target was fixed in advance as the direction with the highest mean positive reuse gain in the original frozen-transfer matrix. Table~\ref{tab:functional_results} reports all 15 conditions that satisfy the source-competence criterion.

\begin{table*}[t]
\caption{Target results for the 15 conditions that satisfy the source-competence criterion. $B$ is original minus controlled-variant target score; values are means $\pm$ sample standard deviations over three seeds.}
\label{tab:functional_results}
\centering\scriptsize
\resizebox{\linewidth}{!}{%
\begin{tabular}{@{}lllrrrc@{}}
\toprule
Source $\rightarrow$ Target & Variant & Metric & Original & Reorganized & $B$ & Seed signs \\
\midrule
Wikipedia $\rightarrow$ Reddit & Sequence & AP & $0.9643\pm0.0028$ & $0.9645\pm0.0026$ & $-0.0002\pm0.0014$ & $+\ +\ -$ \\
Wikipedia $\rightarrow$ Reddit & Time & AP & $0.9643\pm0.0028$ & $0.9646\pm0.0030$ & $-0.0004\pm0.0002$ & $-\ -\ -$ \\
Reddit $\rightarrow$ Wikipedia & Sequence & AP & $0.9286\pm0.0072$ & $0.9262\pm0.0097$ & $+0.0024\pm0.0067$ & $-\ +\ +$ \\
Reddit $\rightarrow$ Wikipedia & Time & AP & $0.9286\pm0.0072$ & $0.9371\pm0.0105$ & $-0.0086\pm0.0108$ & $-\ -\ -$ \\
MOOC $\rightarrow$ Reddit & Partner & AP & $0.8880\pm0.0048$ & $0.8927\pm0.0036$ & $-0.0047\pm0.0066$ & $-\ +\ -$ \\
MOOC $\rightarrow$ Reddit & Sequence & AP & $0.8880\pm0.0048$ & $0.8914\pm0.0058$ & $-0.0034\pm0.0080$ & $-\ +\ -$ \\
MOOC $\rightarrow$ Reddit & Time & AP & $0.8880\pm0.0048$ & $0.8961\pm0.0031$ & $-0.0081\pm0.0068$ & $-\ -\ -$ \\
ICEWS14 $\rightarrow$ ICEWS05-15 & Sequence & MRR & $0.2455\pm0.0384$ & $0.2125\pm0.0052$ & $+0.0330\pm0.0389$ & $+\ +\ +$ \\
ICEWS14 $\rightarrow$ ICEWS05-15 & Time & MRR & $0.2455\pm0.0384$ & $0.2238\pm0.0267$ & $+0.0218\pm0.0321$ & $-\ +\ +$ \\
ICEWS14 $\rightarrow$ ICEWS05-15 & Direction & MRR & $0.2455\pm0.0384$ & $0.2271\pm0.0162$ & $+0.0184\pm0.0385$ & $+\ -\ +$ \\
GDELT $\rightarrow$ Wikipedia & Sequence & AP & $0.9553\pm0.0080$ & $0.9430\pm0.0213$ & $+0.0123\pm0.0262$ & $-\ -\ +$ \\
GDELT $\rightarrow$ Wikipedia & Direction & AP & $0.9553\pm0.0080$ & $0.9536\pm0.0071$ & $+0.0017\pm0.0021$ & $+\ -\ +$ \\
ICEWS05-15 $\rightarrow$ Wikipedia & Sequence & AP & $0.9675\pm0.0031$ & $0.9716\pm0.0015$ & $-0.0041\pm0.0017$ & $-\ -\ -$ \\
ICEWS05-15 $\rightarrow$ Wikipedia & Time & AP & $0.9675\pm0.0031$ & $0.9723\pm0.0022$ & $-0.0048\pm0.0023$ & $-\ -\ -$ \\
ICEWS05-15 $\rightarrow$ Wikipedia & Direction & AP & $0.9675\pm0.0031$ & $0.9713\pm0.0075$ & $-0.0038\pm0.0044$ & $-\ -\ -$ \\
\bottomrule
\end{tabular}}
\end{table*}

\section{Public Dataset Sources}
\label{sec:appendix_data_sources}
\begin{table*}[t]
\caption{Public dataset sources used in the experiments.}
\label{tab:data_sources}
\centering\small
\setlength{\tabcolsep}{4pt}
\renewcommand{\arraystretch}{1.15}
\begin{tabular}{@{}lll@{}}
\toprule
Format & Dataset & Data source \\
\midrule
CTDG & Wikipedia & \url{https://snap.stanford.edu/jodie/wikipedia.csv} \\
CTDG & Reddit & \url{https://snap.stanford.edu/jodie/reddit.csv} \\
CTDG & MOOC & \url{https://snap.stanford.edu/jodie/mooc.csv} \\
CTDG & LastFM & \url{https://snap.stanford.edu/jodie/lastfm.csv} \\
TKG & ICEWS14 & TLogic: \url{https://github.com/liu-yushan/TLogic} (\texttt{data/icews14}) \\
TKG & ICEWS05-15 & TLogic: \url{https://github.com/liu-yushan/TLogic} (\texttt{data/icews0515}) \\
TKG & GDELT & CyGNet: \url{https://github.com/CunchaoZ/CyGNet} (\texttt{data/GDELT}) \\
\bottomrule
\end{tabular}
\end{table*}

\clearpage
\bibliographystyle{IEEEtran}
\bibliography{references}

\end{document}